\documentclass[11pt]{article} 
\usepackage{macros}
\usepackage{setspace}

\begin{document}

\begin{center}

	{\bf{\LARGE{Extracting robust and accurate features via a robust information bottleneck}}}

	\vspace*{.25in}

 	\begin{tabular}{ccccc}
 		{\large{Ankit Pensia}$^*$} & \hspace*{.5in} & {\large{Varun Jog$^\dagger$}} & \hspace*{.5in} & {\large{Po-Ling Loh$^{\ddagger\mathsection}$}}\\
 		{\large{\texttt{ankitp@cs.wisc.edu}}} & \hspace*{.5in} & {\large{\texttt{vjog@ece.wisc.edu}}} & \hspace*{.5in} & {\large{\texttt{ploh@stat.wisc.edu}}}
 			\end{tabular}
 \begin{center}
 Departments of Computer Science$^*$, Electrical \& Computer Engineering$^\dagger$, and Statistics$^\ddagger$\\
 University of Wisconsin-Madison\\
Department of Statistics, Columbia University$^\mathsection$
 \end{center}

	\vspace*{.2in}

October 2019

	\vspace*{.2in}

\end{center}

\begin{abstract}
We propose a novel strategy for extracting features in supervised learning that can be used to construct a classifier which is more robust to small perturbations in the input space. Our method builds upon the idea of the information bottleneck by introducing an additional penalty term that encourages the Fisher information of the extracted features to be small, when parametrized by the inputs. By tuning the regularization parameter, we can explicitly trade off the opposing desiderata of robustness and accuracy when constructing a classifier. We derive the optimal solution to the robust information bottleneck when the inputs and outputs are jointly Gaussian, proving that the optimally robust features are also jointly Gaussian in that setting. Furthermore, we propose a method for optimizing a variational bound on the robust information bottleneck objective in general settings using stochastic gradient descent, which may be implemented efficiently in neural networks. Our experimental results for synthetic and real data sets show that the proposed feature extraction method indeed produces classifiers with increased robustness to perturbations.
\end{abstract}

\section{Introduction}

Over the past decade, deep learning algorithms have revolutionized modern machine learning, achieving superhuman performance in several diverse scenarios such as image classification~\cite{KriEtal12}, machine translation~\cite{BahEtal14}, and strategy games~\cite{SilEtal16}. These algorithms are distinguished by their ability to solve complex problems by processing massive data sets efficiently with the help of large-scale computing power. On the other hand, as deep learning algorithms are gradually adopted in high-stakes applications such as autonomous driving, disease diagnosis, and legal analytics, it has become increasingly important to ensure that such algorithms are interpretable~\cite{DosKim17}, fair~\cite{BarEtal17}, and secure~\cite{AbaEtal16}. In particular, the lack of ``robustness'' of neural networks, explained in more detail below, has become a significant concern.

It was observed in Szegedy et al.~\cite{SzeEtal13} that the high accuracy of trained neural networks may be compromised under small (nearly imperceptible) changes in the inputs~\cite{EngEtal17}.  Perhaps more alarmingly, empirical studies suggest the existence of certain ``universal adversarial perturbations" that can thwart any neural network architecture~\cite{MooEtal17}. Following these observations, the research area of \emph{robust machine learning} has seen tremendous activity in recent years. Briefly stated, research in robust machine learning considers various threat models and proposes strategies to attack and defend neural networks. More recently, there has been a growing emphasis on certifiable defenses; i.e., defenses that are provably robust against \emph{all} possible adversaries. We briefly describe some relevant work below.

One popular method for increasing the robustness of neural networks is known as data augmentation~\cite{SimEtal03, KriEtal12, RajEtal19}, in which a training data set is enlarged using artificial training points constructed with small perturbations of the inputs. Some authors~\cite{GooEtal14, MadEtal17} suggest augmenting the data set by carefully chosen perturbation directions that approximate worst-case perturbations, the latter of which are  infeasible to compute exactly in high dimensions. Another approach is to smooth the decision boundaries of a trained neural network using a preprocessing step such as randomized smoothing~\cite{LecEtal18, CohEtal19, ZhaLia19}, which may lead to computable certificates on the robustness of a neural network classifier~\cite{RagEtal18, WenEtal18, LiEtal18, CohEtal19, SalEtal19}.
On the other hand, many of the methods for defending against adversarial attacks which initially showed promise have subsequently been broken~\cite{AthEtal18}.

Recent work suggests that high accuracy and high robustness may in fact be in conflict with each other~\cite{GilEtal18, TsiEtal18}, which may even be a fundamental defect of any classifier~\cite{FawEtal18, Doh18, ZhaEtal19, Fel19}. This has suggested certain tradeoffs between maximally robust and maximally accurate classification: If it is desirable to train a classifier which is robust to small perturbations in the inputs, it may be necessary to forego the level of accuracy obtained when such a restriction is not present. Indeed, although randomized smoothing approaches  guarantee the robustness of classifiers in the sense that they are relatively insensitive to perturbations in the input space, more aggressive smoothing (hence, improved robustness) directly compromises the accuracy of the smoothed classifier on clean data~\cite{CohEtal19}.

Tsipras et al.~\cite{TsiEtal18} and Ilyas et al.~\cite{IlyEtal19} have suggested a dichotomy between ``robust" and ``non-robust" features. Although both types of features might be useful for classification---in the sense that the features are positively correlated with the class labels---an adversary may act on the non-robust features to make them no longer relevant for classification. One approach for building a robust classifier would therefore be to train a classifier which only operates on the robust features. In a similar spirit, Garg et al.~\cite{GarEtal18} developed a method for extracting provably robust features in supervised learning problems based on extracting the eigenvectors of the Laplacian of a graph constructed on the training data set. However, this method of robust feature extraction notably disregards any information about the ability of the features to predict the correct classes; instead, the features are first extracted and then a classifier is trained using those features. In contrast, the characterization of robust features proposed by Ilyas et al.~\cite{IlyEtal19} is defined with respect to correlations with the classes.

Inspired by these lines of work and the counterexamples that have emerged to demonstrate non-robustness of classifiers, we propose a new method for extracting robust features. Importantly, our goal is to extract features from data that are at once robust and useful for predicting class labels. However, we quantify usefulness not only in terms of correlations as in Ilyas et al.~\cite{IlyEtal19}, but also in terms of nonlinear information theoretic functionals such as mutual information. Furthermore, we characterize robustness in terms of an appropriately defined notion of Fisher information. Our method is heavily inspired by the information bottleneck objective of Tishby et al.~\cite{TisEtal00}, which we will review in detail; however, whereas the information bottleneck formulation is designed to extract features that are both concise and relevant for prediction, we further augment the objective function to encourage robustness in the constructed features. The bottleneck formulation also allows us to explicitly trade off between robustness and accuracy, reminiscent of the work of Zhang et al.~\cite{ZhaEtal19}, who propose a different regularizer to promote robustness at the cost of accuracy. However, our regularizer is motivated from an information and estimation theoretic point of view, in contrast to their more geometric point of view. Although not directly related, we also point to the work of Achille and Soatto~\cite{AchSoa18, AchSoa19}, where both mutual information and Fisher information have been proposed as useful metrics for measuring the degree to which the weights in a network ``memorize" the training dataset. The difference between Tishby et al.~\cite{TisEtal00} and our work on the one hand, and Achille and Soatto~\cite{AchSoa18, AchSoa19} on the other hand, is that the former considers information content in the activations, whereas the latter considers information content in the weights of a neural network.

The remainder of the paper is organized as follows: In Section~\ref{SecRIB}, we review the information bottleneck methodology and introduce the versions of the robust information bottleneck objective that will be studied in this paper. In Section~\ref{SecProperties}, we derive properties of the proposed Fisher information regularizer, which encourages robustness of the extracted features. In Section~\ref{SecGaussian}, we rigorously derive solutions to the robust information bottleneck objective when the inputs and outputs are jointly Gaussian, and interpret the results. In Section~\ref{SecVariational}, we present a variational optimization framework for obtaining approximate solutions in the case of general distributions. We provide simulation results on synthetic and real data sets in Section~\ref{SecExperiments}, and conclude with a discussion in Section~\ref{SecDiscussion}.

\paragraph{\textbf{Notation:}} Random variables will be denoted by capital letters ($X, Y, Z$), their support will be denoted by calligraphic letters ($\cX, \cY, \cZ$), and their densities will be denoted via subscripts ($p_X, p_Y, p_Z$). Random vectors will be written as column vectors, and when $X = (X_1, \dots, X_n)^\top $ and $Y = (Y_1, \dots, Y_m)^\top $, we will denote $(X, Y) = (X^\top , Y^\top )^\top $. For a vector $v \in \real^p$, we write $v^{\downarrow}$ to denote the $p$-dimensional vector with components rearranged in decreasing order. For two vectors $v,w \in \real^p$, we write $v^{\downarrow} \preceq w^{\downarrow}$ to indicate that $w^{\downarrow}$ majorizes $v^{\downarrow}$, meaning that for all $1 \le k \le p$, we have
\begin{equation*}
\sum_{i=1}^k (v^{\downarrow})_i \le \sum_{i=1}^k (w^{\downarrow})_i.
\end{equation*}
We use the shorthand $[n]$ to denote the set $\{1, \dots, n\}$.

For a matrix $A \in \real^{p \times p}$, let $\lambda(A)$ denote the (multi)set of eigenvalues of $A$. Let $\lambda_{\min}(A)$ and $\lambda_{\max}(A)$ denote the minimum and maximum eigenvalues, respectively. Let $\|A\|_F$ denote the Frobenius norm. Let $\diag(a_1, \dots, a_p)$ denote the $p \times p$ diagonal matrix with $(a_1, \dots, a_p)$ on the diagonal. We write $I_d$ to denote the $d \times d$ identity matrix. In the linear algebraic statements throughout the paper, we will generally consider the singular value decomposition (SVD) to be the ``thin SVD." We write $\Cov(X)$ to denote the covariance matrix of a random vector $X$; and when $Y$ is another random vector, we write $\Cov(X,Y)$ to denote the covariance matrix of the concatenated vector $(X,Y)$. We write $\Cov(X|Y)$ to denote the average conditional covariance matrix of $X$, where the integral is taken with respect to the density of $Y$. We will denote the entropy of a discrete random variable $X$ by $H(X)$, and the differential entropy of a continuous random variable $X$ by $H(X)$, as well.

\section{Problem formulation}
\label{SecRIB}

Consider a dataset $(X,Y) \sim p_{XY}$, where $X$ is thought of as a sample corresponding to a label $Y$. The information bottleneck theory proposed in Tishby et al.~\cite{TisEtal00} is a variational principle used for extracting as much relevant information about $Y$ from $X$ as possible, while achieving the largest possible compression of $X$. Using mutual information to measure ``relevance'' and ``compression,'' Tishby et al.~\cite{TisEtal00} proposed the optimization problem
\begin{equation}
    \inf_{p_{T|X}( \cdot \mid \cdot)} \left\{I(T; X) - \gamma I(T; Y)\right\}.
\end{equation}
The extracted feature, denoted by $T$, is a random function of $X$ generated by the kernel $p_{T|X}$. Since it does not directly depend on $Y$, we have the Markov chain $Y \to X \to T$. The parameter $\gamma > 0$ trades off compression and relevance of the extracted features $T$. Over the years, the information bottleneck principle has found myriad applications in learning problems \cite{SloEtal00, CheEtal03, GonEtal03, ChaEtal16, AleEtal16}. More recently, information bottleneck theory has also been used to gain insight into the training of deep neural networks. By measuring the information content of different layers in a network, it was observed that layers in a neural network undergo two separate phases, one consisting of a \emph{memorization phase} where both $I(T;X)$ and $I(T;Y)$ increase, and a \emph{compression phase} where $I(T;X)$ decreases while $I(T;Y)$ continues to increase~\cite{TisEtal15, ShaTis17}.

Broadly speaking, a ``bottleneck'' formulation tries to trade off between two quantities; in the information bottleneck formulation, these quantities are relevance and compression, each measured using mutual information. In this paper, we seek a formulation that trades off relevance and robustness. Depending on the specific learning problem under consideration, one may measure relevance and robustness using variety of metrics. We present two natural formulations below. %

\subsection{Measuring relevance}

In the information bottleneck formulation, relevance is captured by the term $I(Y;T)$. Apart from mutual information being a natural quantity to consider, we may justify $I(Y;T)$ in another way: Suppose the support of $Y$ is a discrete set $\cY$ such that $|\cY| = 2^M < \infty$, and  $Y$ has a uniform distribution over $\cY$. (This is natural if $Y$ corresponds to the label of an image $X$.) If $I(Y;T) \geq M - \epsilon$, then equivalently, $H(Y|T) \leq \epsilon$. Thus, intuitively, there is little uncertainty about $Y$ after observing $T$. In practical settings, we may consider the MAP estimate $\hat Y (T)$. It has been shown \cite[Theorem 1]{FedMer94} that the probability of error for the MAP estimate may be upper-bounded by a function of $H(Y|T)$, so a small value of $H(Y|T)$ implies a high accuracy of the MAP decoder.

Notably, if $\cY$ is a continuous space, it is not possible in general to bound the error of the MAP estimate in terms of $H(Y|T)$. Moreover, in such cases, a squared-error metric may be more relevant than the probability of error. To accommodate such scenarios, we propose to measure relevance using $\mmse(Y|T)$, which denotes the minimum MSE for a predictor of $Y$ constructed using $T$. Equivalently, we may write
\begin{equation*}
\mmse(Y|T) = \E[(Y - \E(Y|T))^2] = \tr\left(\Cov(Y|T)\right).
\end{equation*}

\subsection{Measuring robustness}

Intuitively, a feature $T$ is robust if small perturbations in $X$ do not change the distribution of $T$ significantly. We may think of the distribution of $T$ as being parametrized by $X$. The sensitivity (being the opposite of robustness) of $T$ to $X$ may then be measured using the (statistical) Fisher information $\Phi(T | X)$, given below: 
\begin{equation*}
\Phi(T|X) = \int_{\cX} \left( \int_{\cT} \left\|\nabla_x \log p_{T|X}(t|x)\right\|_2^2 p_{T|X}(t|x) dt \right) p_X(x) dx := \int_{\cX} \Phi(T | X = x) p_X(x) dx.
\end{equation*}
The quantity $\Phi(T|X=x)$ may be interpreted as the sensitivity of the distribution of $p_{T|X}(\cdot | x)$ to changes at the point $x$. Integrating over $\cX$, we see that $\Phi(T|X)$ measures the average sensitivity of the log conditional density with respect to changes in the inputs $X$. It can be shown that under  mild regularity conditions on the densities of $X$ and $T$, we have (cf.\ Appendix~\ref{AppFisher})
\begin{equation*}
\Phi(T|X) = J(X|T) - J(X),
\end{equation*}
where $J(X) = \E\left[\|\nabla_x \log p_X(x)\|_2^2\right]$ and $J(X|T) = \int p_T(t) \left(\int \|\nabla_x \log p_{X|T} (x|t) \|_2^2 p_{x|t}(x|t) dx \right) dt$. The quantity $J(\cdot)$ is often called the information theorist's Fisher information, which is different from the statistical Fisher information $\Phi(\cdot | \cdot)$.

Naturally, Fisher information is not the only measure of robustness (or sensitivity) one may use. As we will show in Section~\ref{SecProperties}, however, the Fisher information satisfies several properties which make it an attractive measure of sensitivity. %

\subsection{Robust information bottleneck objective}

Since we want to extract features that are simultaneously relevant and robust, we define the features determined by the robust information bottleneck to be the optimum of
\begin{equation}
\label{EqnMMSEFish}
\inf_{p_{T|X}(\cdot | \cdot)} \left\{\mmse(Y|T) + \beta \Phi(T|X)\right\},
\end{equation}
or
\begin{equation}
\label{EqnIBFish}
\inf_{p_{T|X}(\cdot | \cdot)} \left\{-I(T;Y) + \gamma I(T;X) + \beta \Phi(T|X)\right\},
\end{equation}
depending on what notion of ``relevance'' is being employed. The latter formulation simply adds the Fisher information term as a regularizer to the canonical information bottleneck.

The operators considered in the above formulations, namely Fisher information, mutual information, and mmse, are all invariant to changes in the scale (or indeed, any smooth bijective transformation) of $T$. The invariance under bijective transformations of $T$ is critical---it would be unnatural to expect an extracted feature to become more (or less) robust by simply taking functions of that feature. The following lemma makes this statement more precise. The proof is contained in Appendix~\ref{AppLemInvariance}. Additional important properties of the Fisher information in relation to robustness are explored in Section~\ref{SecProperties}.

\begin{lemma*}
\label{lemma: invariance}
Let $Y \to X \to T$ be a Markov chain, such that $T$ is an $\real^d$-valued random vector. Let $f: \real^d \to \real^d$ be a smooth bijection. Then the following equalities hold:
\begin{enumerate}
\item
$I(X;T) = I(X; f(T))$ and $I(Y;T) = I(Y; f(T))$.
\item
$\mmse(Y|T) = \mmse(Y|f(T))$.
\item
$\Phi(T|X) = \Phi(f(T)|X)$.
\item
If $T = (T_1, T_2)$ is such that $T_2 \ind (T_1, X, Y)$, then $I(X;T) = I(X; T_1)$, $I(Y; T) = I(Y; T_1)$, $\mmse(Y|T) = \mmse(Y|T_1)$, and $\Phi(T|X) = \Phi(T_1|X)$. In other words, the independent component $T_2$ may be ignored when characterizing the optimal solution to the robust information bottleneck. 
\end{enumerate}
\end{lemma*}

 Further note that the standard information bottleneck formulation is invariant not only to transformations of $T$, but also to transformations of $X$ and $Y$. However, this is not the case for the robust information bottleneck formulation, since $\Phi(T|X) \neq \Phi(T|f(X))$ in general. This is another attractive property of the robust information bottleneck formulation: If data are pre-processed so that the distribution $p_X(\cdot)$ is squeezed along a certain direction---for example, by multiplying $X$ by a diagonal matrix $\diag(1,1, \dots, 1, \epsilon)$---the robustness with respect to perturbations along the final dimension should be reduced in comparison to the other directions. The standard information bottleneck formulation is {blind} to such transformations and extracts the same features regardless of transformations of $X$, whereas the robust information bottleneck {adapts to the scaling} of $p_X(\cdot)$.

We end with a lemma showing that analogous to formulation \eqref{EqnMMSEFish}, it is possible to consider a modified version of \eqref{EqnIBFish} where we drop the $I(X;T)$ term. Specifically, consider the optimization problem

\begin{equation}
\label{EqnMIFish}
\inf_{p_{T|X}(\cdot | \cdot)} \left\{-I(T;Y) + \beta \Phi(T|X)\right\}.
\end{equation}
A concern with this formulation could be that the value of $I(X;T)$ may be arbitrarily large at the optimum of formulation \eqref{EqnMIFish}, leading to features that are not concise, although they may be robust. Our next lemma, proved in Appendix~\ref{AppLemPhiMI}, shows that this cannot happen and that robustness also implies compression:

\begin{lemma*}
\label{lemma: phi and mi}
Let $X \sim p_X$ be an $\real^p$-valued random variable, and let $T$ be an extracted feature via the channel $p_{T|X}$. Then following inequality holds:
\begin{equation}
I(X;T) \leq H(X) - \frac{p}{2} \log \frac{2\pi e p}{\Phi(T|X) + J(X)}.
\end{equation}
In particular, if $\Phi(T|X)$ is bounded from above, then so is $I(X;T)$.
\end{lemma*}

\subsection{Examples}

Before proceeding further, we describe two examples in the case when the input distribution is a Gaussian mixture. We will illustrate the instantiation of the Fisher information term as a regularizer, and return to these examples in the simulations to follow in Section~\ref{section: Gaussian mixture}.

Suppose $Y$ takes values $+ \mathbf 1 := (1, 1)^\top $ and $- \mathbf 1 := (-1, -1)^\top$, with probability $1/2$ each. Conditioned on $Y = +\mathbf 1$, the distribution of $X$ is $\cN(+\mathbf 1, \text{Diag}(\sigma_1^2, \sigma_2^2))$; and conditioned on $Y = -\mathbf 1$, the distribution of $X$ is $\cN(-\mathbf 1, \text{Diag}(\sigma_1^2, \sigma_2^2))$. (This is identical to an example considered in Ilyas et al. \cite{IlyEtal19}.)

\begin{example}
\label{ExaLinear}
In the first setting of interest, we will consider random features $T$ parametrized by $w := (w_1, w_2)^\top  \in \real^2$ as 
$T = w^\top  X + \xi,$ where $\xi \sim \cN(0, 1)$.
\end{example}

\begin{example}
\label{ExaLogistic}
We will also consider a setting where $T$ is a binary feature taking values $\pm 1$, following a logistic distribution with parameter $w$:
\begin{equation*}
\mprob(T = 1 | X = x) = \frac{1}{1+\exp(-x^\top w)}.
\end{equation*}
\end{example}

The main questions we wish to investigate are:
How does the optimal $w$ change with the regularization parameter $\beta$? How does the optimal $w$ change with $\sigma_1^2$ and $\sigma_2^2$? We shall do this by evaluating the terms in our proposed robust information bottleneck objective, and then solving the joint optimization problem using numerical techniques. 

The following two lemmas derive convenient closed-form expressions for the Fisher information, without making any assumptions on the distribution of $X$. However, we will use them to analyze the settings of Examples~\ref{ExaLinear} and~\ref{ExaLogistic}, respectively, when $X$ follows a Gaussian mixture, in which case it will be simpler to assess the quality of the extracted features.

\begin{lemma*}
\label{LemFeatGauss}
Suppose $T = AX + \epsilon$, where $\epsilon \sim N(0, I)$. Then
\begin{equation*}
\Phi(T|X) = \|A\|_F^2,
\end{equation*}
so adding a Fisher information penalty is in this case equivalent to $\ell_2$-regularization.
\end{lemma*}

As Lemma~\ref{LemFeatGauss} shows, the Fisher information directly encodes the signal-to-noise ratio (SNR) of the channel from $X$ to $T$. If the SNR is low, small changes in $X$ have less of an effect on the distribution of $T$, meaning the features are more robust. In addition to quantifying this insight, Lemma~\ref{LemFeatGauss} will be useful for our calculations later. The proof is contained in Appendix~\ref{AppLemFeatGauss}.

Now suppose that we instead extract a binary feature. The proof of the following lemma is contained in Appendix~\ref{AppLemFeatBin}.

\begin{lemma*}
\label{LemFeatBin}
Suppose $T \in \{+1, -1\}$ is a binary feature such that
\begin{equation*}
\mprob(T = 1 \mid X = x) = \frac{1}{1 + \exp(-x^\top  w)}.
\end{equation*}
Then
\begin{equation*}
\Phi(T|X = x) =  \|w\|_2^2 \cdot \mprob(T = 1 \mid X = x) \cdot \mprob(T = -1 \mid X = x).
\end{equation*}
\end{lemma*}

The empirical approximation to $\Phi(T|X) = \int \Phi(T|X = x) p_X(x) dx$ will be
\begin{equation}
\label{EqnNoisyLogistic}
\frac{1}{n} \sum_{i=1}^n \Phi(T|X = x_i) = \frac{\|w\|_2^2}{n} \sum_{i=1}^n \mprob(T = 1 | X = x_i) \cdot \mprob(T = -1 | X = x_i).
\end{equation}
We see from the formula in Lemma~\ref{LemFeatBin} that the Fisher penalty encourages more confident predictions. At the same time, the norm $\|w\|_2$ is encouraged to be small, relating to the discussion of the SNR following Lemma~\ref{LemFeatGauss}. In Section~\ref{section: Gaussian mixture}, we detail experiments that show how the Fisher penalty indeed encourages adversarial robustness.

\begin{remark*}
The expression~\eqref{EqnNoisyLogistic} has shown up in Wager et al.~\cite{WagEtal13} as a ``quadratic noising penalty," which is a first-order approximation of a regularizer obtained by adding noise to inputs when performing maximum likelihood estimation in logistic regression. On the other hand, a key difference in our setting is that the conditional probabilities appearing in the expression are for the \emph{feature} $T$ conditioned on $X = x_i$, whereas the setting of Wager et al.~\cite{WagEtal13} involves the probabilities of $Y$ conditioned on $X = x_i$. Indeed, although artificially adding noise to inputs (i.e., data augmentation) has been empirically shown to improve robustness in certain situations, the goal of our approach is to explicitly extract robust features, which are then used to predict the correct class labels. Furthermore, we can treat the term $\Phi(T|X)$ as a regularizer to be used in conjunction with any loss (e.g., MMSE or mutual information-based), whereas the derivations in Wager et al.~\cite{WagEtal13} are specific to the maximum likelihood loss for logistic regression.

Note also that Wager et al.~\cite{WagEtal13} mention a regularizer involving a Fisher information term in their analysis of dropout, but this has no analogy to our setting because the Fisher information there is computed with respect to the parameters involved in prediction, whereas the Fisher information here is computed by taking gradients with respect to the inputs themselves.
\end{remark*}

\section{Robustness properties of Fisher information}
\label{SecProperties}

One of our motivations in using the Fisher information as a proxy for sensitivity is its amenability to analysis. Indeed, the Fisher information is a well-studied quantity in both information theory and estimation theory~\cite{GuoEtal13}. Another motivation is our interest in the accuracy versus robustness tradeoff and its connection to the Cram\'{e}r-Rao inequality from statistics~\cite{Leh06}. The Cram\'{e}r-Rao inequality (or its generalization, the van Trees inequality) states that for a parameter $\Theta \sim p_{\Theta}$ and a family of distributions $p_{X|\Theta}$, the following inequality holds:
\begin{align*}
\mmse(\Theta|X) \geq \frac{1}{\Phi(X|\Theta) + J(\Theta)}.
\end{align*}
In other words, high robustness (low $\Phi(X|\Theta)$) leads to lower accuracy (a larger lower bound on $\mmse(\Theta|X)$). Apart from these motivations, we will now prove several additional statements concerning Fisher information that make it a suitable proxy for sensitivity.

Having extracted robust features $T$, we first note that \emph{any} classifier that uses $T$ to predict $Y$ is guaranteed to be robust, as well. This supports the observation of Ilyas et al.~\cite{IlyEtal19}, who show empirically that classifiers trained using ``robust'' features are also robust. The proof of Lemma~\ref{LemYhat} is contained in Appendix~\ref{AppLemYhat}.

\begin{lemma*}
\label{LemYhat}
Let $Y \to X \to T \to \widehat Y$ be a Markov chain. Here, we think of $T$ as an extracted feature and $\widehat Y$ as a prediction of $Y$ using $T$. The sensitivity of $\widehat Y$ to perturbations in $X$ is measured by $\Phi(\widehat Y | X)$. Then the following bound holds:
\begin{equation*}
\Phi(\widehat Y | X) \leq \Phi(T|X).
\end{equation*}
In other words, the output $\widehat Y$ is at least as robust as the extracted features $T$.
\end{lemma*}

The next two lemmas relate the Fisher information to small changes in mutual information and KL divergence, thus showing that an upper bound on $\Phi(T|X)$ guarantees lower sensitivity of the respective quantities. The following lemma, proved in Appendix~\ref{AppLemGaussPerturb}, provides an interpretation of the term $\Phi(T|X)$ in terms of regularizing the effect of small perturbations to the mutual information:

\begin{lemma*}
\label{LemGaussPerturb}
Suppose $Z \sim N(0,I)$ is a standard normal random variable that is independent of $(X, Y, T)$. Then
\begin{equation*}
I(X; T) - I(X+ \sqrt \delta Z; T) =  \frac{\delta}{2} \Phi(T|X) + o(\delta).
\end{equation*}
\end{lemma*}

Lemma~\ref{LemGaussPerturb} shows that adding the Fisher information term $\Phi(T|X)$ encourages the mutual information between $X$ and $T$ to only change slightly under small Gaussian perturbations. However, as seen in the simulations, this also leads to more robust classifiers in terms of adversarial perturbations.

Another way to interpret the Fisher information term is as follows: Let $\epsilon > 0$ and let $u$ be a unit vector. An extracted feature $T$ will be considered robust for a particular $X = x$ if the distributions $p_{T|X}(\cdot | X = x)$ and $p_{T|X}(\cdot | X = x + \epsilon u)$ are not too different for any choice of $u$ and all small enough $\epsilon$. The difference between these two distributions could be measured by a number of metrics, but we focus on the KL divergence here. Note that the KL divergence provides an upper bound on the total variation distance, and also bounds Wasserstein distances in certain special cases \cite{RagSas13}. (Wasserstein distance is the metric of study in recent work on distributional robustness~\cite{SinEtal17, StaJeg17}; however, the goal of such studies is to directly learn neural network models that are distributionally robust to the inputs, rather than our intermediate step of extracting robust features.)

The proof of the following result is contained in Appendix~\ref{AppLemKLPerturb}:

\begin{lemma*}
\label{LemKLPerturb}
Let $\|u\|_2 = 1$. Let $x  + \epsilon u$ be a small perturbation of $x$ in the direction $u$. Then we have
\begin{equation*}
D(p_{T|X = x+ \epsilon u} \| p_{T|X = x}) = \frac{\epsilon^2}{2} \Phi(T | X = x) + o(\epsilon^2).
\end{equation*}
\end{lemma*}

Since the right-hand expression does not depend on the direction $u$, Lemma~\ref{LemKLPerturb} shows that when $x$ is perturbed arbitrarily in a ball of radius $\epsilon$, the corresponding distribution of $T$ lies in a KL-ball of radius $\frac{\epsilon^2}{2} \Phi(T|X=x)$ around the distribution $p_{T|X=x}(\cdot | X = x)$. Requiring $\Phi(T|X = x)$ to be small on average is equivalent to requiring $\Phi(T|X)$ to be small, so adding this term as a penalty encourages the algorithm to extract features that are robust to arbitrary $\ell_2$-perturbations, on average. Note that this is identical to the objective of adversarial training in Madry et al.~\cite{MadEtal17}.

Finally, we show that the upper bound on the KL divergence in Lemma~\ref{LemKLPerturb} can be translated into a direct guarantee on robustness. Consider a (deterministic) classifier $g: \mathcal{T} \rightarrow \mathcal{Y}$ that maps extracted features to a predicted label, and a classifier $f: \mathcal{X} \rightarrow \mathcal{Y}$ defined by
\begin{equation*}
f(x) := \arg\max_{y} p_{T|X = x} (g(t) = y).
\end{equation*}
In practice, we could approximate the value of $f$ by generating random features according to the distribution $T|X = x$, applying the map $g$, and taking the majority vote over the result.

The main idea, which is motivated by an argument found in Zhang and Liang~\cite{ZhaLia19},  is to use the fact that an upper bound on the KL divergence between distributions implies an upper bound on total variation distance. Hence, if we have an input $x \in \mathcal{X}$ such that the \emph{classification margin}, defined by
\begin{equation*}
\margin_f(x,y) := p_{T|X = x} (g(t) = y) - \max_{z \neq y} p_{T|X = x} (g(t) = z),
\end{equation*}
is sufficiently large, then we should also have $f(x') = f(x)$ when $x'$ is contained in a small ball around $x$.

For the result to follow, we assume that the $o(\epsilon^2)$ bound on the remainder in Lemma~\ref{LemKLPerturb} is uniform over all choices of $x$, which holds if the third-degree differential of $p_{T|X = x}$ with respect to $x$ is uniformly bounded. The proof is provided in Appendix~\ref{AppLemRobBound}.

\begin{lemma*}
\label{LemRobBound}
For any $\epsilon, \eta > 0$, we have
\begin{equation}
\label{EqnRobBound}
\mprob\Big(f(x') = f(x) \quad \forall x' \in B_\epsilon(x)\Big) \ge \mprob(x \in B^\eta) - \frac{\epsilon^2 \Phi(T|X) + o(\epsilon^2)}{\eta},
\end{equation}
where
\begin{align*}
B^\eta & := \left\{x \in \mathcal{X}: \margin_f(x,f(x)) > \sqrt{\eta}\right\}.
\end{align*}
\end{lemma*}

The expression on the right-hand side of inequality~\eqref{EqnRobBound} provides a lower bound on the probability that a randomly chosen input is robust to perturbations of magnitude bounded by $\epsilon$ in any direction. Furthermore, the lower bound is higher when $\mprob(x \in B^\eta)$ is larger; i.e., the distribution on $\mathcal{X}$ is such that a larger fraction of points have high margin. To further interpret Lemma~\ref{LemRobBound}, suppose the distributions of $X$ and $T|X$ are fixed, and consider the effect of adjusting the parameters $\epsilon$ or $\eta$. Note that if we increase $\epsilon$, the ball $B_\epsilon(x)$ in which the classifier is guaranteed to be robust becomes larger; however, the right-hand side of inequality~\eqref{EqnRobBound} decreases, leading to a weaker probabilistic guarantee. On the other hand, if we decrease $\eta$ to increase the probability $\mprob(x \in B^\eta)$ appearing in the lower bound, then the term $\frac{\epsilon^2 \Phi(T|X) + o(\epsilon^2)}{\eta}$ also increases. Thus, we see that tradeoffs exist in determining the optimal choices of both $\epsilon$ and $\eta$.

\section{Jointly Gaussian variables}
\label{SecGaussian}

In general, it is impossible to obtain closed-form expressions for the solutions to the optimization problems~\eqref{EqnMMSEFish} and~\eqref{EqnIBFish}. However, as in the case of the canonical information bottleneck, the optimization problems become more tractable when $(X,Y)$ have a jointly Gaussian distribution~\cite{CheEtal05}. In this section, we derive explicit formulas for the solutions to the optimization problems in order to develop some theoretical intuition for the similarities and differences between the robust information bottleneck formulations, and to verify that the extracted features are in fact meaningful in special cases. We will assume throughout this section that $\Sigma_x \succ 0$ and $\Cov(Y|X) = \Sigma_y - \Sigma_{yx} \Sigma_x^{-1} \Sigma_{xy} \succ 0$.

\subsection{Information bottleneck formulation}
\label{SubSecIB}

We first study the information bottleneck formulation~\eqref{EqnIBFish}. Our proof uses a technique for establishing information inequalities pioneered by Geng and Nair \cite{GengNair14}. Geng and Nair showed that it is enough to establish certain \emph{subadditivity} relations for functionals in order to establish Gaussian optimality; this strategy has been used to prove a variety of entropy and information inequalities in the past few years \cite{CouJia14, KimEtal16, Gol16,  YangEtal17, ZhaEtal18, Cou18, AnaEtal19}. The proof of optimality is provided in detail in Appendix~\ref{AppIBOptimality}, and we only provide a proof sketch here.

\subsubsection{Optimality}

Let $(X_G,Y_G)$ be jointly Gaussian random variables. As before, we express $Y_G = CX_G + \xi$, where $\xi$ is independent of $X$. We may rewrite the robust information bottleneck formulation as
\begin{multline}
\sup_{p_{T|X_G}(\cdot | \cdot)} \left\{I(T;Y_G) - \gamma I(T; X_G) - \beta \Phi(T|X_G)\right\} \\
=  \left[ \sup_{p_{T|X_G}(\cdot | \cdot)} \Big\{- H(Y_G|T) + \gamma H(X_G|T) - \beta J(X_G|T)\Big\}\right] + H(Y_G) - \gamma H(X_G) + \beta J(X_G) \label{eq: square}.
\end{multline}
Since we are only concerned with the optimizing $p_{T|X_G}$, we shall focus on the optimization problem in the square brackets. Consider the function $f$ defined on the space of densities $p_X$ over $\real^p$:
\begin{align*}
f(X) = -H(CX+\xi) + \gamma H(X) - \beta J(X) := -H(Y) + \gamma H(X) - \beta J(X),
\end{align*}
where we use $Y := CX + \xi$ to indicate the output channel that scales the input by $C$ and adds Gaussian noise $\xi$ to the scaled input. The upper-concave envelope of $f$ is given by
\begin{align*}
F(X) = \sup_{p_{T|X}(\cdot | \cdot)} \left\{- H(Y|T) + \gamma H(X|T) - \beta J(X|T)\right\},
\end{align*}
where we allow $\abs \cT$ to be countably large for now. Note that the optimization problem in the square brackets in equation~\eqref{eq: square} is equivalent to finding the optimizing $T$ in the upper-concave envelope of $f$ at the particular distribution $X_G$. We shall use use the subadditivity proof strategy from Geng and Nair \cite{GengNair14}. To do so, we define a lifting of $f$ to pairs of random variables, as follows:
\begin{align*}
f(X_1, X_2) = -H(Y_1, Y_2) + \gamma H(X_1, X_2) - \beta J(X_1, X_2).
\end{align*}
Let $F(X_1, X_2)$ be the upper-concave envelope of $f(X_1, X_2)$. Our main result is the following subadditivity lemma:

\begin{lemma*}\label{lemma: subaddinf}
For any pair of random variables $(X_1, X_2)$, we have 
\begin{align*}
F(X_1, X_2) \leq F(X_1) + F(X_2).
\end{align*}
\end{lemma*}

\begin{proof}
Note that
\begin{align*}
f(X_1, X_2 | T) &:= -H(Y_1, Y_2 | T) + \gamma H(X_1, X_2 | T) - \beta J(X_1, X_2 | T)\\
&\stackrel{(a)}= \left(-H(Y_1|T) + \gamma H(X_1 | T) - \beta J(X_1 | X_2, T)\right) \\
& \qquad \qquad + \left(-H(Y_2|Y_1, T) + \gamma H(X_2 |X_1, T) - \beta J(X_2 | X_1, T) \right)\\
&\stackrel{(b)}\leq \left(-H(Y_1|T) + \gamma H(X_1 | T) - \beta J(X_1 | T)\right) \\
& \qquad \qquad + \left(-H(Y_2|X_1, T) + \gamma H(X_2 |X_1, T) - \beta J(X_2 | X_1, T) \right)\\
&= f(X_1 | T) + f(X_2 | X_1, T)\\
&\stackrel{(c)}\leq F(X_1) + F(X_2).
\end{align*}
Here, $(a)$ follows from the chain rule of Fisher information in Lemma~\ref{Lem:JPhi}, $(b)$ follows from the data processing inequality for Fisher information in Lemma~\ref{Lem:JConvex}, and $(c)$ follows from the definition of $F$.
\end{proof}
Our next lemma is obtained by following the proof technique pioneered in Geng and Nair~\cite{GengNair14}. The proof of this lemma is lengthy, but the techniques employed are becoming relatively standard in the information theory literature. We have provided a detailed proof in Appendix~\ref{AppIBOptimality}. 

\begin{lemma*}
Consider the optimization problem
\begin{align*}
V(K) := \sup_{\Cov(X) \preceq K} f(X).
\end{align*}
The optimizer of the above problem is a unique Gaussian random variable $X^* \sim \cN(0, K^*)$, with $K^* \preceq K$. In particular, $f(X^*) = F(X^*) = V(K)$. 
\end{lemma*}

Returning to the robust information bottleneck formulation for jointly Gaussian $(X_G, Y_G) \sim p_{X_GY_G}$, let $\Cov(X_G) = K$. Let $X^* \sim \cN(0, K^*)$ be the optimizer that achieves $V(K)$. Let $X' \perp \!\!\! \perp X$ be such that $X' \sim \cN(0, K-K^*)$, so $X^*+X'$ has the same distribution as $X_G$. It is easy to check that
\begin{align*}
F(X_G) \geq f(X_G|X') = f(X^*) = V(K).
\end{align*}
However, we also have
\begin{align*}
F(X_G) \leq F(X^*) = V(K),
\end{align*}
where the first inequality comes from the fact that $X^*$ maximizes both $f$ and $F$. This shows that the optimal joint distribution $(T,X_G)$ may be taken to be $(X', X_G)$; i.e., $T = X'$. Since $(X', X_G)$ are jointly Gaussian, this proves that it is enough to consider random variables $T$ that are jointly Gaussian with $X_G$ to solve the optimization problem~\eqref{EqnIBFish}. Note that the joint distribution of $(X_G, T)$ has covariance 
$$\begin{pmatrix}
K &K - K^*\\
K - K^* &K - K^*
\end{pmatrix}.$$
Thus, we may write $T = DX_G + N$, where 
\begin{align*}
D &= (K-K^*)K^{-1}, \quad \text{ and }\\
N &\sim \cN(0, (K-K^*) - (K-K^*)K^{-1}(K-K^*)).
\end{align*}
Since the scaling of $T$ does not matter, we can also rewrite the optimizing $T$ as $T = \tilde D X + \tilde N$, where
\begin{align}
\tilde D &= \left[(K-K^*) - (K-K^*)K^{-1}(K-K^*)\right]^{-1/2} (K-K^*)K^{-1}, \quad \text{ and } \nonumber \\
\tilde N &\sim \cN(0, I). \label{EqnT}
\end{align}

\subsubsection{Identity covariance}

We now derive an explicit form of the optimal feature map in the case when $\Sigma_x$ is a multiple of the identity. The proof of the following theorem is contained in Appendix~\ref{AppThmInfoId}.

\begin{theorem*}
\label{ThmInfoId}
Suppose $\Sigma_x = \sigma_x^2 I$. Let $B = (\Sigma_y - \Sigma_{yx} \Sigma_x^{-1} \Sigma_{xy})^{-1/2} \Sigma_{yx} \Sigma_x^{-1}$, and let $B = V \Lambda W^\top $ be the SVD. Let $\Lambda = \diag(\ell_1, \dots, \ell_k)$, where the diagonal elements are sorted in decreasing order. For each $i \le k$, define
\begin{equation}
\label{EqnUniv}
d_i = \arg\min_{d \in [0,1]} \left\{\frac{1}{2} \log\left(\frac{\sigma_x^2 d}{\ell_i} + 1\right) - \frac{\gamma}{2} \log d + \frac{\beta}{\sigma_x^2 d}\right\},
\end{equation}
and let $D = \diag(d_1, \dots, d_k)$.
Let $\hat{U}$ be the permutation matrix which sorts the diagonal entries of $D$ in increasing order, and let $U = W \hat{U}^\top $. The optimal feature map is then given by $T = \frac{1}{\sigma_x} (D^{-1} - I)^{1/2} U^\top  X + \epsilon$, where $\epsilon \sim N(0, I_k)$.
\end{theorem*}

To summarize, the optimal projection directions are given by a permutation/rearrangement of the right singular vectors $W$ appearing in the SVD of $B$, together with appropriate rescalings obtained by optimizing the univariate functions~\eqref{EqnUniv}. As will be described in further detail in Section~\ref{SecComp}, this resembles the solution to the usual information bottleneck.

\begin{remark*}
Note that as stated in Theorem~\ref{ThmInfoId}, we can always find an optimal feature map into $k$ dimensions, where $k = \rank(B)$. On the other hand, it is possible that the optimal feature map could be expressible in even fewer dimensions, e.g., if some of the $d_i$'s are equal to 1.
\end{remark*}

\subsubsection{The case $\gamma = 0$}
\label{SecMutual}

Now suppose we instead consider the mutual information formulation
\begin{align}
\label{EqnMutualFish}
\min_{p_{T|X}(\cdot | \cdot)} \left\{I(Y;T) + \beta \Phi(T|X)\right\},
\end{align}
which is a special case of the formulation~\eqref{EqnIBFish} when $\gamma = 0$. The benefit of this formulation is that we can obtain a closed-form solution without assuming an identity covariance, under the condition that $\beta$ is not too large. Furthermore, note that by Lemma~\ref{lemma: phi and mi}, an upper bound on $\Phi(T|X)$ also implies an upper bound on $I(X;T)$, so setting $\gamma = 0$ in the robust information bottleneck objective should still lead to reasonable results in terms of both robustness and relevance when $\beta$ is chosen sufficiently large.

We have the following result, proved in Appendix~\ref{AppThmGamZero}:

\begin{theorem*}
\label{ThmGamZero}
Suppose $\beta$ is sufficiently small. Let $C = \Sigma_x^{-1/2} \Sigma_{xy} (\Sigma_y - \Sigma_{yx} \Sigma_x^{-1} \Sigma_{xy})^{-1/2}$ and suppose $\Sigma_{xy}$ has full column rank. Consider the SVDs
\begin{align*}
C & = W \Lambda V^\top , \\
(C^\top  \Sigma_x^{-1} C)^{-1} & = UDU^\top .
\end{align*}
Define $\tilde{D} = \diag(\tilde{d}_1, \dots, \tilde{d}_k)$ to be a diagonal matrix with
\begin{equation*}
\tilde{d}_i = \frac{1 + \sqrt{1 + 4d_i/\beta}}{2d_i/\beta},
\end{equation*}
where $D = \diag(d_1, \dots, d_k)$. Let $S\Gamma S^\top $ be the SVD of $\Lambda V^\top  U \tilde{D}^{-1} U^\top  V \Lambda - I$.
The optimal feature map is given by $T = \Gamma^{1/2} S^\top  W^\top  \Sigma_x^{-1/2} X + \epsilon$, where $\epsilon \sim N(0,I_k)$.
\end{theorem*}

As seen in the proof of the theorem, the required upper bound on $\beta$ can be expressed in terms of the spectra of $(\Sigma_x, \Sigma_y, \Sigma_{xy})$.

\subsection{MMSE formulation}

We now consider the MMSE formulation~\eqref{EqnMMSEFish}. As in the previous section, we will provide a proof sketch for optimality that may be converted into a rigorous proof by following the steps in Appendix~\ref{AppIBOptimality}. Moreover, we will derive the optimal form of $T$ when $(X_G, Y_G)$ are jointly Gaussian, expressed in terms of their associated covariance matrices.

\subsubsection{Optimality}
\label{SecOptMMSE}

 Using the chain rule for Fisher information in Appendix~\ref{AppFisher}, we can derive the following lemma:

\begin{lemma*}
\label{lemma: condition}
The optimization objective in formulation~\eqref{EqnMMSEFish} can be equivalently expressed as
\begin{multline*}
\min_{p_{T|X}(\cdot | \cdot)} \left\{ \mmse(Y|T) + \beta \Phi(T|X) \right\} = \tr(\Cov(Y)) + \beta J(X) - \max_{p_{T|X}(\cdot | \cdot)} \left\{ \tr(\Cov (Y|T)) - \beta J(X|T)\right\}.
\end{multline*}
\end{lemma*}

\begin{proof}
Note that
\begin{align*}
\mmse(Y|T) &= \tr \Big(\Cov(Y) - \Cov(Y|T) \Big) \quad \text{ and }\\
\Phi(T|X) &= J(X|T) - J(X),
\end{align*}
where the second equation follows from Lemma~\ref{Lem:JPhi}. Since the joint distribution $p_{XY}$ is fixed, we may remove the $\tr(\Cov(Y))$ and $J(X)$ terms from the optimization objective and arrive at the desired result.
\end{proof}

Since $X_G$ and $Y_G$ are jointly Gaussian, we can always express $Y_G$ as
\begin{equation*}
Y_G = CX_G + \xi,
\end{equation*}
where $C = \Sigma_{yx} \Sigma_x^{-1}$ and $\xi \perp \!\!\! \perp X$ is a Gaussian random variable with $\Cov(\xi) = \Sigma_y - \Sigma_{yx} \Sigma_x^{-1} \Sigma_{yx}$.
Consider the following function defined on the space of densities $p_X$ over $\cX$:
\begin{align*}
f(X) = \tr(\Cov(CX+\xi)) - \beta J(X).
\end{align*}
With this definition, it is easy to see that 
\begin{align*}
F(X) := \sup_{p_{T|X}(\cdot | \cdot)} \left\{\tr(\Cov (CX+\xi|T)) - \beta J(X|T)\right\}
\end{align*}
is simply the upper-concave envelope of $f(\cdot)$. Define a lifting of $F$ to pairs of random variables as
\begin{align*}
F(X_1, X_2) = \max_{p_{T|X_1, X_2}(\cdot | \cdot)} \left\{\tr(\Cov (CX_1 + \xi_1, CX_2 + \xi_2|T)) - \beta J(X_1, X_2|T)\right\},
\end{align*}
where $\xi_1$ and $\xi_2$ are independent Gaussian random variables with the same marginal distribution as $\xi$. Furthermore, the optimization is taken over any conditional density $p_{T|X_1, X_2}$, where we assume that the joint distribution of $(X_1, X_2)$ is fixed.

The main step is to establish a subadditivity lemma, analogous to Lemma~\ref{lemma: subaddinf}. We show the following:
\begin{lemma*}
\label{lemma: subaddmmse}
The function $F$ is subadditive, i.e., 
\begin{align*}
F(X_1, X_2) \leq F(X_1) + F(X_2).
\end{align*}
\end{lemma*}

\begin{proof}
Let $Y_1 = CX_1 + \xi_1$ and $Y_2 = CX_2 + \xi_2$. Then we have
\begin{align*}
\tr(\Cov(Y_1, Y_2|T)) = \tr(\Cov(Y_1|T)) + \tr(\Cov(Y_2|T)).
\end{align*}
Furthermore, the chain rule for Fisher information gives
\begin{align*}
J(X_1, X_2|T) = J(X_1 | X_2, T) + J(X_2 | X_1, T).
\end{align*}
Hence, we have
\begin{align*}
f(X_1, X_2 | T) &= \tr(\Cov(Y_1, Y_2 | T)) - \beta J(X_1, X_2 | T) \\
&= \tr(\Cov(Y_1|T)) - \beta J(X_1 | X_2, T) + \tr(\Cov(Y_2|T))  -\beta J(X_2 | X_1, T)\\
&\leq \tr(\Cov(Y_1| T)) - \beta J(X_1 |  T) + \tr(\Cov(Y_2| T))  - \beta J(X_2 |  T)\\
&= f(X_1 | T) + f(X_2 | T)\\
&\leq F(X_1) + F(X_2),
\end{align*}
where the first inequality uses the data processing property of the covariance and Fisher information operators. Optimizing over $T$, we conclude the desired subadditivity result.
\end{proof}
We shall not prove the lemma below, but the proof follows the steps outlined in Geng and Nair~\cite{GengNair14}, and also in our Appendix~\ref{AppIBOptimality}:

\begin{lemma*}
Consider the optimization problem
\begin{align*}
V(K) := \sup_{\Cov(X) \preceq K} f(X).
\end{align*}
Then the optimizer of the above problem is a unique Gaussian random variable $X^* \sim \cN(0, K^*)$ with $K^* \preceq K$. In particular, $f(X^*) = F(X^*) = V(K)$.
\end{lemma*}

Let $\Cov(X_G) = K$. Let $X^*$ be the optimizer that achieves $V(K)$. Let $X' \perp \!\!\! \perp X$ be such $X' \sim \cN(0, K-K^*)$, so that $X^*+X'$ has the same distribution as $X$. Following an identical argument as in Section~\ref{SubSecIB}, we can show that it is enough to consider $(X,T) \sim (X, X')$, and thus, $T$ is jointly Gaussian with $X_G$.

Again, as in Section~\ref{SubSecIB}, we may write $T = DX_G + N$, where 
\begin{align*}
D &= (K-K^*)K^{-1}, \quad \text{ and }\\
N &\sim \cN(0, (K-K^*) - (K-K^*)K^{-1}(K-K^*)).
\end{align*}
Since the scaling of $T$ does not matter, we can also rewrite the optimizing $T$ as $T = \tilde D X + \tilde N$, where
\begin{align*}
\tilde D &= \left[(K-K^*) - (K-K^*)K^{-1}(K-K^*)\right]^{-1/2} (K-K^*)K^{-1}, \quad \text{ and } \nonumber \\
\tilde N &\sim \cN(0, I).
\end{align*}

\subsubsection{Identity covariance}
\label{SecIdentity}

The following theorem derives a closed-form expression for the optimal feature map in the case when $\Sigma_x$ is a multiple of the identity. The proof is contained in Appendix~\ref{AppThmMMSEId}.

\begin{theorem*}
\label{ThmMMSEId}
Suppose $\Sigma_x = \sigma_x^2 I$. Let $0 < \lambda_1 \le \cdots \le \lambda_k$ denote the ordered nonzero eigenvalues of $\Sigma_{xy} \Sigma_{yx}$. Define $U \in \real^{p \times k}$ to be the matrix with columns equal to the ordered unit eigenvectors corresponding to $(\lambda_1, \dots, \lambda_k)$. For $1 \le i \le k$, define
\begin{equation*}
d_i =
\begin{cases}
\sqrt{\frac{\lambda_i}{\beta}} - 1 & \text{ if } \lambda_i \ge \beta, \\
0 & \text{otherwise,}
\end{cases}
\end{equation*}
and let $D = \diag(d_1, \dots, d_k)$. Then the optimum choice of features is given by $T = AX + \epsilon$, where $A = \frac{1}{\sigma_x} D^{1/2} U^\top $ and $\epsilon \sim N(0,I_k)$.
\end{theorem*}

\begin{remark*}
\label{RemGaussGen}
When $\Sigma_x \neq \sigma_x^2 I$, it is difficult to obtain a closed-form solution for the optimization problem~\eqref{EqnMMSEFish}. However, as in the variational approach detailed in Section~\ref{SecVariational} for extracting features in non-Gaussian settings, it may be instructive to minimize an upper bound on the objective function.

In particular, using the fact that the trace of a product of psd matrices is upper-bounded by the product of traces, we have
\begin{equation}
\label{EqnTraceBd}
\tr\left(D \cdot \beta U^\top  \Sigma_x^{-1} U\right) \le \beta \tr(D) \tr(U^\top  \Sigma_x^{-1} U) = \beta \tr(D) \tr(\Sigma_x^{-1}).
\end{equation}
Hence, we can replace the intermediary optimization problem~\eqref{EqnGenObj} appearing in the derivation by the surrogate problem
\begin{equation}
\label{EqnSurrObj}
\min_{D,U} \left\{\tr\left((D+I)^{-1} \cdot U^\top  \Sigma_x^{-1/2} \Sigma_{xy} \Sigma_{yx} \Sigma_x^{-1/2} U\right) + \beta \tr(D) \tr(\Sigma_x^{-1})\right\},
\end{equation}
where the surrogate objective function upper-bounds the original objective function.

Importantly, using the same logic as in the proof of Theorem~\ref{ThmMMSEId}, the optimal choice of $U$ does not depend on $D$. In this case, we can argue that the columns of $U$ correspond to the (ordered) eigenvectors of $\Sigma_x^{-1/2} \Sigma_{xy} \Sigma_{yx} \Sigma_x^{-1/2}$. Optimizing with respect to $D$ then corresponds to solving
\begin{equation*}
\min_{d_1, \dots, d_p} \sum_{i=1}^p \left(\frac{1}{d_i + 1} \cdot \lambda_i + \beta \tr(\Sigma_x^{-1} d_i\right),
\end{equation*}
so that
\begin{equation*}
d_i =
\begin{cases}
\sqrt{\frac{\lambda_i}{\beta \tr(\Sigma_x^{-1})}} - 1 & \text{ if } \lambda_i \ge \beta \tr(\Sigma_x^{-1}), \\
0 & \text{otherwise.}
\end{cases}
\end{equation*}
The optimal projection will then be $A = D^{1/2} U^\top  \Sigma_x^{-1/2}$.

Note that when $\Sigma_x = \sigma_x^2 I$, the inequality~\eqref{EqnTraceBd} is actually an equality. Consequently, the optimum of the surrogate objective exactly agrees with the optimum derived in Theorem~\ref{ThmMMSEId} in that case.
\end{remark*}

\subsection{Comparison between solutions}
\label{SecComp}

Now that we have derived explicit formulae for the optimal feature maps in several settings (Theorems~\ref{ThmInfoId}, \ref{ThmGamZero}, and~\ref{ThmMMSEId}), it is instructive to compare the form of the solutions. In particular, we see that all of the feature maps may be expressed as $T = AX + \epsilon$, with $\epsilon \sim N(0, I_k)$, where $A = \tilde{D} \tilde{U}^\top  \Sigma_x^{-1/2}$. In each case, $\tilde{D} \in \real^{k \times k}$ is an appropriate diagonal matrix and $\tilde{U} \in \real^{p \times k}$ is a matrix with $k$ orthonormal columns, taken from the spectral decomposition of some matrix function of $(\Sigma_x, \Sigma_y, \Sigma_{xy})$.

Digging a bit deeper, we see that the scaling matrix $\tilde{D}$ will generally depend critically on the value of the regularization parameter $\beta$. In particular, as $\beta \rightarrow \infty$, successive entries of $\tilde{D}$ will be truncated to 0 (e.g., $d_i \rightarrow 1$ in Theorem~\ref{ThmInfoId} and $d_i \rightarrow 0$ in Theorem~\ref{ThmMMSEId}). This same behavior is manifest in the canonical information bottleneck formulation for jointly Gaussian variables (cf.\ Theorem 3.1 of Chechik et al.~\cite{CheEtal05}). The transition points are accordingly referred to as ``critical points" for $\beta$. In our formulation, where the regularization parameter $\beta$ trades off robustness and accuracy, it is natural that larger values of $\beta$ will lead to zeroing out features (which are then very robust but completely useless in prediction); at the other extreme, small values of $\beta$ lead to a full feature map which preserves all eigenvectors, regardless of the magnitude of the corresponding eigenvalues.

Turning to the eigenvector matrix $\tilde{U}$, we see that the particular matrix varies somewhat according to the precise robust bottleneck formulation. Comparing the two identity covariance cases, we see that for the information bottleneck formulation (Theorem~\ref{ThmInfoId}), we are interested in the right singular vectors of $(\Sigma_y - \Sigma_{yx} \Sigma_{xy})^{-1/2} \Sigma_{yx}$. For the MMSE formulation (Theorem~\ref{ThmMMSEId}), we are interested in the eigenvectors of $\Sigma_{xy} \Sigma_{yx}$. The spectra of these matrices will in general be somewhat different, although it is worth noting that when $\Sigma_y$ is a multiple of the identity, the spectrum of $\Sigma_{xy} \Sigma_{yx}$ also furnishes the solution for the canonical information bottleneck formulation and is the relevant spectrum for canonical correlation analysis (CCA). Furthermore, in the special case when $Y$ is 1-dimensional, all of these formulations will result in picking out the same unit vector corresponding to a rescaled version of $\Sigma_{xy}$, agreeing with intuition.

\section{Variational bounds}
\label{SecVariational}

Although our work is motivated by robustness considerations in deep learning, the framework we have developed thus far does not involve any assumptions that the classifier we employ for predicting $Y$ from $T$ is a neural network. In this section, we see how properties of neural networks may be leveraged for the purpose of optimization.

The objective~\eqref{EqnMMSEFish} is intractable to minimize explicitly except in certain special cases. In general, we propose to minimize an appropriate upper bound. Inspired by a recent line of work on variational approximations to the information bottleneck objective~\cite{AleEtal16}, we describe the upper bound and a tractable optimization procedure that uses stochastic gradient descent. Our goal is to learn $p_{T|X}(\cdot|\cdot)$ to solve the optimization problem~\eqref{EqnMMSEFish}.

\begin{remark*}
In this section, we will focus mostly on deriving variational bounds for the objective~\eqref{EqnMMSEFish}, which will involve obtaining appropriate upper bounds for both the MMSE and Fisher information terms separately. Indeed, analogous variational upper bounds for the objective~\eqref{EqnIBFish} may then be obtained by adapting the arguments in Alemi et al.~\cite{AleEtal16}, since the information bottleneck term used for measuring relevance is the same in that paper, and we simply need to add the extra variational expression for the Fisher information term. See Section~\ref{SecAlemi} for several brief remarks.
\end{remark*}

We shall restrict ourselves to kernels $p_{T|X}(\cdot|\cdot)$ that are parametrized by $\theta$. Let $K \in \mathbb N$.  Specifically, we shall consider $p_{T|X}(\cdot | x) = \cN(\mu(x; \theta), \Sigma(x; \theta))$,  where $\mu(\cdot; \theta)$ and $\Sigma(\cdot; \theta)$ are the mean and variance of a Gaussian density that are parametrized by $\theta$. We shall also assume that $\Sigma(x; \theta)$ is a diagonal matrix with entries $\sigma_i^2(x; \theta)$, for $i \in [K]$. For neural networks, the parameters $\theta$ correspond to the weights of a network that takes inputs $x$ and has $2K$ outputs corresponding to $
\mu_i$ and $\sigma^2_i$. These $2K$ outputs will represent $p_{T|X}(\cdot | x)$, as follows: The first $K$ outputs correspond to the means $\mu_i(x)$, and the latter $K$ outputs correspond to variances $\sigma_i^2(x)$. 

Our goal is to derive a variational upper bound on the optimization objective in \eqref{EqnMMSEFish}, so that the upper bound may be maximized efficiently using mini-batch stochastic gradient descent. This will allow us to implement the robust information bottleneck in deep network architectures. The current formulation is not amenable to such techniques, since evaluating $\mmse (Y|T)$ requires knowledge of the intractable posterior distribution $p_{Y|T}$.

\subsection{Optimizing the $ \mmse$ term}

The $\mmse$ has the following variational characterization:
\begin{align*}
\mmse(Y|T) = \inf_{f: \cT \to \cY} \E (Y-f(T))^2,
\end{align*}
where the infimum is achieved by the conditional expectation	function; i.e., 
\begin{align*}
f^*(t) = \E(Y | T = t).
\end{align*}
Calculating $f^*$ requires evaluating the posterior $p_{Y|T}(y|t)$, which we wish to avoid. Thus, we propose to use the upper bound
\begin{align*}
\mmse(Y|T) \leq \E (Y - \tilde f(T))^2,
\end{align*}
for a suitable function $\tilde f: \cT \to \cY$ that is easy to compute. This function $\tilde f$ may be parametrized by some parameters $\phi$, which will be updated during iterations of stochastic gradient descent. (See Section~\ref{SecStochGrad} for details.)

\subsection{Optimizing the Fisher information term}

The term $\Phi(T|X)$ may be efficiently optimized in its original form, and we do not need to derive tractable variational bounds for it. To see this, notice that
\begin{align*}
\Phi(T|X) &= \int_{\cX} \Phi(T | X = x) p_X(x) dx.
\end{align*}
The density $p_{T|X}(\cdot | x)$ is $\cN(\mu(x; \theta), \Sigma(x; \theta))$, and thus,
\begin{align*}
p_{T|X}(t|x) = \frac{1}{\sqrt{(2\pi)^k \prod_{i=1}^k \sigma^2_i(x)}} \exp \left( -\sum_{j=1}^k \frac{(t_j - \mu_j(x))^2}{2\sigma_j(x)^2} \right).
\end{align*}
Taking the logarithm and the gradient with respect to $x$, we obtain
\begin{align*}
\nabla_x  \log p_{T|X}(t|x)  &= \nabla_x \left( -\frac{k}{2} \log 2\pi - \sum_{j=1}^k \log \sigma_j(x)-\sum_{j=1}^k \frac{(t_j - \mu_j(x))^2}{2\sigma_j(x)^2} \right)\\
&= -\sum_{j=1}^k \frac{\nabla_x \sigma_j(x)}{\sigma_j(x)} + \sum_{j=1}^k \frac{(t_j - \mu_j(x))}{\sigma_j(x)^2} \nabla_x \mu_j(x) + \sum_{j=1}^k \frac{(t_j - \mu_j(x))^2}{\sigma_j(x)^3} \nabla_x \sigma_j(x).
\end{align*}
We can express $\Phi(T|X=x)$ as
\begin{align*}
\Phi(T | X= x) &= \int_{\real^k} \| \nabla_x \log p_{T|X}(t|x)\|_2^2 p_{T|X}(t | x) dt\\
&= \int_{\real^k} \Bigg \| - \sum_{j=1}^k \frac{\nabla_x \sigma_j(x)}{\sigma_j(x)} + \sum_{j=1}^k \frac{(t_j - \mu_j(x))}{\sigma_j(x)^2} \nabla_x \mu_j(x) + \sum_{j=1}^k \frac{(t_j - \mu_j(x))^2}{\sigma_j(x)^3} \nabla_x \sigma_j(x) \Bigg\|_2^2\\
&\hspace{5cm} \frac{1}{\sqrt{(2\pi)^k \prod_{j=1}^k \sigma^2_j(x)}} \exp \left( -\sum_{j=1}^k \frac{(t_j - \mu_j(x))^2}{2\sigma_j(x)^2} \right) dt.
\end{align*}

Note that what is essential for us is being able to compute the \emph{derivative} of $\Phi(T|X)$ with respect to the parameters $\theta$ (that parametrize $p_{T|X}$). We now explore how to evaluate the (stochastic) gradients of $\mmse(Y|T)$ and $\Phi(T|X)$ using the reparametrization trick of Kingma and Welling~\cite{KinWel13}.

\subsection{Evaluating stochastic gradients}
\label{SecStochGrad}

Suppose that we have a data set $\{(x_i, y_i): i \in [n]\}$. The empirical distribution is given by 
\begin{align}\label{EqnEmpirical}
\mprob_n(x,y) := \frac{1}{n} \sum_{i=1}^n \delta(x_i, y_i).
\end{align}
As noted earlier, the empirical version of the variational approximation to the $\mmse$ term is given by
\begin{align*}
\E_{\mprob_n}\left[(Y - \tilde f(T))^2\right] = \frac{1}{n} \sum_{i=1}^n \int_{\real^k} (y_i - \tilde f(t; \phi))^2 p_{T|X}(t|x_i; \theta) dt,
\end{align*}
where we have have explicitly included the parameters $\phi$ and $\theta$ to indicate that they parametrize $\tilde f$ and $p_{T|X}$, respectively. Note that stochastic gradient descent involves calculating the gradient of this function in with respect to $\phi$ and $\theta$, so it is critically important to be able to evaluate these derivatives in a computationally feasible manner. We use the reparametrization trick from Kingma and Welling~\cite{KinWel13} to write $T = \tau(X, \epsilon; \theta)$, where $\epsilon$ is a standard normal variable that is independent of $X$, and $\tau$ is a function that is parametrized by $\theta$.

\subsection{Details about the reparametrization}

We have expressed $p_{T|X = x}$ as the Gaussian density $\cN(\mu(x; \theta), \Sigma(x; \theta))$. Alternatively, 
\begin{align*}
T = \mu(x; \theta) + \Sigma(x; \theta)^{1/2} \epsilon =: \tau(x, \epsilon; \theta),
\end{align*}
 where $\epsilon \sim \cN(0, I)$.

Rewriting the MMSE integral, we have
\begin{align*}
\frac{1}{n} \sum_{i=1}^n \int_{\real^k} (y_i - \tilde f(t; \phi))^2 p_{T|X}(t|x_i; \theta) dt &= \frac{1}{n} \sum_{i=1}^n \int_{\real^k} (y_i - \tilde f(\tau(x_i, \epsilon; \theta); \phi))^2 p_{\epsilon}(\epsilon) d\epsilon.
\end{align*}
Furthermore, we may approximate the integral over $\epsilon$ by resampling $\epsilon_1, \dots, \epsilon_{mn}$ from the distribution $p_\epsilon$, and computing
\begin{align*}
\frac{1}{nm} \sum_{i=1}^n \sum_{j=1}^m (y_i - \tilde f(\tau(x_i, \epsilon_{ij}; \theta); \phi))^2 \epsilon_{ij}.
\end{align*}

Finally, note that the gradient of this function with respect to $\theta$ (or $\phi$) may be calculated easily using backpropagation, since we may use the trained neural networks to evaluate the functions $\tau_\theta$ and $\tilde f_\phi$, as well as the gradients of these functions with respect to either their parameters or their inputs. This shows how to take the gradient of the $\mmse$ term.

We now examine the Fisher information term. Using the reparametrization above, we may write
\begin{align}
\Phi(T | X) & \approx \E_{\mprob_n}[\Phi(T|X = x)] \nonumber\\
& = \frac{1}{n}\sum_{i=1}^n \int_{\real^k} \Bigg \|-\sum_{j=1}^k \frac{\nabla_x \sigma_j(x_i)}{\sigma_j(x_i)} + \sum_{j=1}^K \frac{(\tau(x_i, \epsilon; \theta) - \mu_j(x_i))}{\sigma_j(x_i)^2} \nabla_x \mu_j(x_i) \nonumber \\
& \qquad \qquad \qquad + \sum_{j=1}^k \frac{(\tau(x_i, \epsilon; \theta) - \mu_j(x_i))^2}{\sigma_j(x_i)^3} \nabla_x \sigma_j(x_i) \Bigg\|_2^2 p_\epsilon(\epsilon) d\epsilon. 
\label{EqnPhiVarGauss}
\end{align}
Again, we may approximate the gradient by sampling from the distribution $p_\epsilon$ and then computing stochastic gradients with respect to $\theta$ using backpropagation. Note that this will require us to calculate expressions such as $\nabla_\theta \nabla_x \mu_j(x)$ and $\nabla_\theta \nabla_x \sigma_j(x)$, which may be computationally intensive depending on the dimension of $x$, but can still be obtained from the trained neural network classifier $\tau_\theta$. Putting together the results, we conclude that the expression~\eqref{EqnMMSEFish} has an upper bound that may approximated and optimized using mini-batch based stochastic gradient descent.

\subsection{Bounds on the mutual information terms}
\label{SecAlemi}

In this section, we briefly discuss the variational bounds for $I(X;T)$ and $I(Y;T)$ in Alemi et al.~\cite{AleEtal16}. Looking at formulation~\eqref{EqnIBFish}, we see that we need an upper bound on $I(X;T)$ and a lower bound on $I(Y;T)$. Let $\widehat Y$ be the estimate of $Y$ based on $T$. This means that we have the Markov chain $Y \to X \to T \to \widehat Y$. The lower bound on $I(Y;T)$ is given by
\begin{align*}
I(Y;T) \geq  \int p_{XY}(x,y) p_{T|X}(t|x) \log p_{\widehat Y|T}(y|t) dx dy dt.
\end{align*}
For the empirical data distribution~\eqref{EqnEmpirical}, this bound evaluates to
\begin{align*}
\frac{1}{N}\sum_{i=1}^N \int p_{T|X}(t|x_i) \log p_{\widehat Y | T} (y_i | t) dt.
\end{align*}
In other words, the variational approximation to $I(Y;T)$ is essentially the cross-entropy loss. 

Turning to $I(X;T)$, Alemi et al.~\cite{AleEtal16} show that
\begin{align*}
I(X;T) \leq  \int p_X(x) p_{T|X}(t|x) \log \frac{p_{T|X}(t|x)}{p_Z(t)} dx dt,
\end{align*}
where $Z \sim p_Z(\cdot)$ is an arbitrary random variable, but is taken to be a standard normal variable in practice. For the empirical distribution~\eqref{EqnEmpirical}, this bound evaluates to
\begin{align*}
I(X;T) &\leq \frac{1}{N}\sum_{i=1}^N \int p_{T|X}(t|x_i) \log \frac{p_{T|X}(t|x_i)}{p_Z(t)} dx dt\\
&= \frac{1}{N}\sum_{i=1}^N D(p_{T|X = x_i} \| p_Z).
\end{align*}
Viewed this way, the variational approximation to $I(X;T)$ is a regularizer that encourages the learned kernels $p_{T|X}(\cdot | x_i)$ to be spherically symmetric. The variational approximation also has the additional feature of introducing a scale in the optimization problem, since the classical information bottleneck formulation is invariant to scaling of $T$. 

As noted earlier, the $\Phi(T|X)$ term can be interpreted as a regularizer that enforces smoothness conditions on the mapping $x_i \to p_{T|X=x_i}$. In summary, the variational approximation to formulation~\eqref{EqnIBFish} consists of three terms: (a) cross-entropy loss; (b) a regularizer that encourages spherical symmetry of learned representations and enforces a scaling constraint; and (c) a regularizer that encourages smoothness. If we eliminate (b) from the optimization procedure, as we propose in formulation~\eqref{EqnMIFish}, a concern would be that there is no scaling constraint in the problem. This may potentially lead to learnt parameters that become arbitrarily large, which may lead to non-convergence during training. However, we have observed in practice that eliminating the regularizer $(b)$ does not have any notable detrimental effects on the learned representations.

\section{Experiments}
\label{SecExperiments}

We now provide simulation results showing the behavior of the various feature extraction methods we have proposed. We begin with a variety of experiments involving synthetic data generated from a Gaussian mixture. We then provide experiments on MNIST data.

\subsection{Gaussian mixture}\label{section: Gaussian mixture}

\begin{figure}[H]
\centering
\scalebox{1.2}{\input{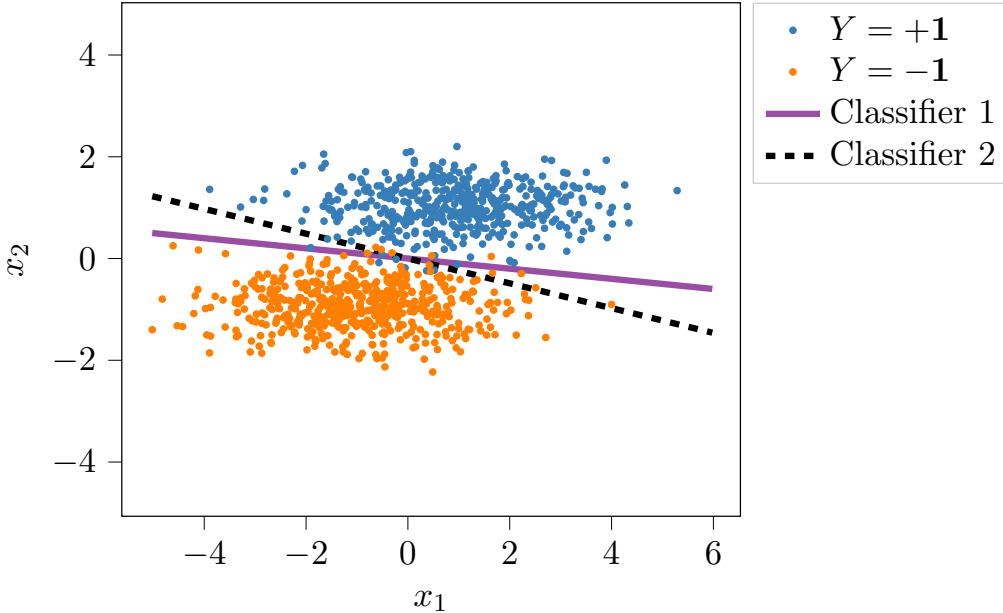}}
\caption{Plot showing point clouds of $1000$ samples for Example~\ref{ExaLinear}, with $\sigma_1^2 = 2$ and $\sigma_2^2 = 0.2$.
The class $Y=+\mathbf{1}$ is centered at $(1,1)$ and the class $Y = -\mathbf{1}$ is centered at $(-1,-1)$.
}
\label{FigExaLinear}
 \end{figure}

We begin by conducting simulations for the setting described in Example~\ref{ExaLinear}. Figure~\ref{FigExaLinear} shows point clouds of $1000$ points for the case of $\sigma_1^2 = 2$ and $\sigma_2^2 = 0.2$. 
Note that in the absence of adversarial perturbations, the decision boundary of the optimal classifier should be close to the horizontal axis (Classifier $1$ in Figure~\ref{FigExaLinear}). Equivalently, the angle of $w^*$ should be close to $90^{\circ}$. 
However, this classifier will not be optimal if we require robustness to adversarial $\ell_2$-perturbations:
Since the decision boundary is close to $x_1$-axis, it is easy for the adversary to perturb the $x_2$-coordinate of a data point and cause the classifier to make an error, since a large number of points are \textit{near} the boundary of Classifier $1$.
Thus, a robust classifier should tilt the boundary slightly to protect against $\ell_2$-perturbations, leading to Classifier $2$.
This intuition is formalized in equation~\eqref{EqAdvAccuracy} and Figure~\ref{FigAdvAccAngle} below, where we can see how the robustness of a classifier varies as the angle of the linear classifier tilts.

We will now show that imposing a robustness constraint by adding a Fisher information $\Phi(T|X)$ encourages a similar effect. By Lemma~\ref{LemFeatGauss}, we have $\Phi(T|X) = \|w\|_2^2$. Turning to the MMSE term, for a given $w$, we denote the features $T$ by $T_w$.  First, we note that conditioned on $Y = +\mathbf 1$ and $-\mathbf 1$, the distribution of $T_w$ is $\cN(w_1 + w_2, w_1^2 \sigma_1^2 + w_2^2 \sigma_2^2 + 1)$ and $\cN(-w_1 - w_2, w_1^2 \sigma_1^2 + w_2^2 \sigma_2^2 + 1)$, respectively. Let $\mu_w := w_1 + w_2$ and $\sigma^2_w := w_1^2 \sigma_1^2 + w_2^2 \sigma_2^2 + 1$. We see that
\begin{align*}
\prob(Y = + \mathbf 1 | T_w = t) = \frac{\exp\left(-\frac{(t-\mu_w)^2}{2\sigma_w^2}\right)}{\exp\left(-\frac{(t-\mu_w)^2}{2\sigma_w^2}\right) + \exp\left(-\frac{(t+\mu_w)^2}{2\sigma_w^2}\right)} = \frac{1}{1 + \exp\left(- \frac{2\mu_w t}{\sigma_w^2} \right)} := \alpha_{t,w}.
\end{align*}
Also,
\begin{equation*}
\mprob(Y = -\mathbf 1|T_w=t) = 1 - \alpha_{t,w} := \bar{\alpha}_{t,w}.
\end{equation*}

The minimum mean squared error for a fixed $w$ then equals
\begin{align*}
\mmse(Y|T_w) & = \frac{1}{2} \int 2\left(1-(\alpha_{t,w} - \bar{\alpha}_{t,w})\right)^2 p(t|Y = + \mathbf 1) dt \\
& \qquad + \frac{1}{2} \int 2\left(-1 - (\alpha_{t,w} - \bar{\alpha}_{t,w})\right)^2 p(t|Y = - \mathbf 1) dt \\
& = \int_\real \frac{8 \alpha_{t,w}^2}{\sqrt{2 \pi \sigma_w^2}}  e^{- \frac{(t+\mu_w)^2}{2\sigma_w^2}}dt = f\left( \frac{\mu_w}{\sigma_w}  \right),
\end{align*}
where $f(\cdot)$ is defined as follows:
\begin{align*}
 f(a) =  \E_{Z \sim \mathcal{N}(a^2, a^2) } \frac{8}{\left(1 + e^{2Z}\right)^2}.
\end{align*}
Thus, the MMSE of $w$ depends only on the scalar $\frac{\mu_w^2}{ \sigma_w^2 }$. Monte Carlo approximation shows that $f(\cdot)$ is a decreasing function on positive reals.
 
Combining the two calculations, the optimal $w_*$ solves the following optimization problem:
\begin{align}
 \arg\min_{w} \qquad & f\left( \frac{\mu_w}{\sigma_w}\right) &\equiv && \arg\max_{w} \qquad & \frac{\mu_w}{\sigma_w} \nonumber \\
s.t. \qquad& \|w\|_2 \leq R & && s.t. \qquad& \|w\|_2 \leq R.
 \label{EqnGMMOpt}
 \end{align}
Note that the optimal value in equation~\eqref{EqnGMMOpt} is achieved at $\|w\|_2 = R$.  
A smaller value of $R$ corresponds to increased robustness in features.

We briefly describe the performance of linear classifiers in the presence of an adversary.
Consider linear classifiers of the form $\text{sign}(w^\top x)$. For such classifiers, adversarial accuracy in the presence of $\epsilon$-corruption in the $\ell_2$-metric is given by
\begin{align}
\epsilon\text{-Adversarial-Accuracy} = \mathbb P\left\{Z \geq  \frac{\epsilon \|w\|_1 - \mu_w}{ \sqrt{w^\top  \Sigma w}} \right\},
\label{EqAdvAccuracy}
\end{align}
where $\epsilon=0$ corresponds to the accuracy of the classifier in the absence of any adversary, and $Z$ is a standard normal random variable. It follows that the performance of such classifiers depends on $w$ only through its direction.
We parametrize the direction by $\theta$, the angle between $w$ and the horizontal axis, measured counter-clockwise.
As the level of perturbation $\epsilon$ changes, the optimal $\theta^*$ changes considerably. Figure~\ref{FigAdvAccAngle} shows the relationship between $\epsilon$-Adversarial-Accuracy and the classifier angle for different values of $\epsilon$. From the figure, we can see that the choice of classifier depends crucially on the desired level of robustness.

\begin{figure}[H]
\centering
\scalebox{1.15}{\input{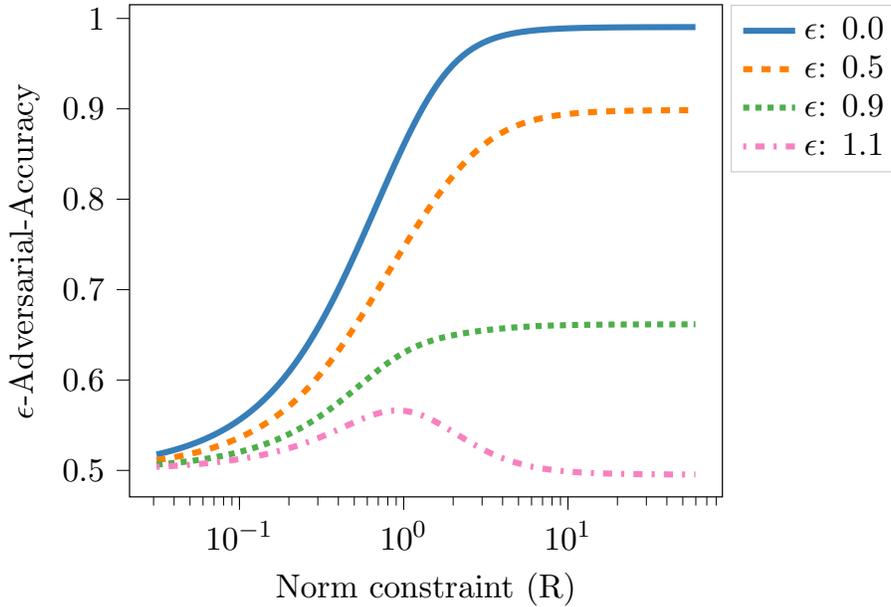}}
\caption{Plot showing the adversarial accuracy for linear classifiers as a function of the norm constraint $R$ and $\epsilon$, the maximum perturbation allowed in the $\ell_2$-metric. For each $R$, we solve the optimization problem in equation~\eqref{EqnGMMOpt} and then calculate its adversarial accuracy for different $\epsilon$. We take $\sigma_1^2 = 2$ and $\sigma_2^2 = 0.2$.}
\label{FigFisherAdvAcc}
 \end{figure}

We now show that the same phenomenon occurs for the classifier obtained by solving equation~\eqref{EqnGMMOpt}.
Note that as $R\to 0$ in equation~\eqref{EqnGMMOpt}, the direction of the optimal $w^*$ tends to $45^\circ$.
As $R \to 0$, the extracted feature $w^\top x + \xi$ corresponds to independent Gaussian noise, so we expect $50\%$ adversarial accuracy for any $\epsilon$.
  Figure~\ref{FigFisherAdvAcc} shows the relation between the norm constraint $R$ and the $\epsilon$-Adversarial-Accuracy for several values of $\epsilon$. 
  As expected, as $R \to 0$, the extracted feature becomes independent of $X$ and the accuracy tends to $50\%$.
  For $\epsilon=0$, i.e., in absence of any adversarial perturbation, the accuracy of the classifier degrades monotonically as we constrain the norm to be smaller.
  The curve corresponding to $\epsilon=1.1$ is more insightful, showing that the performance of the classifier increases at first as we constrain $\|w\|_2$ to be smaller. If we further increase the constraint (making $R$ smaller), the extracted feature tends towards Gaussian noise and the performance degrades.

\begin{figure}[H]
\centering
\scalebox{1.15}{\input{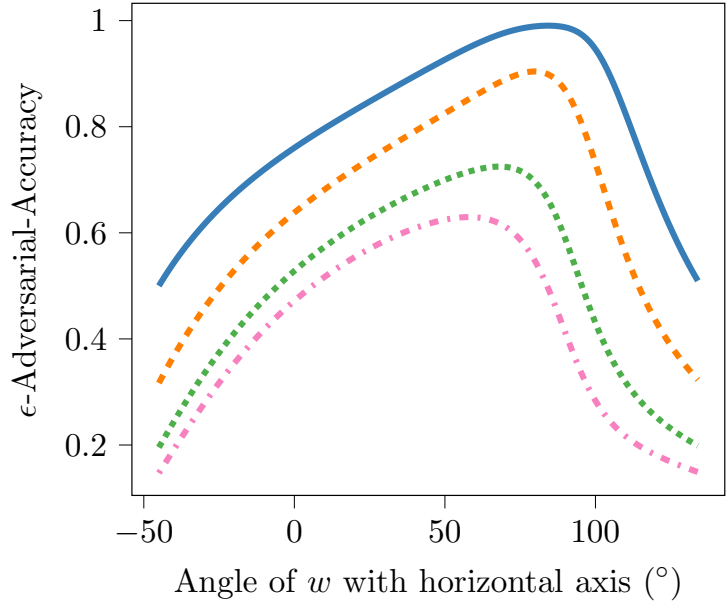}}
\caption{Plot showing the adversarial accuracy for linear classifiers, $\text{sign}(w^\top x)$,  as a function of the angle of $w$ and $\epsilon$, the maximum perturbation allowed in the $\ell_2$-metric. The adversarial accuracy of such classifiers depends on $w$ only through its direction, which we parametrize by the angle of $w$ with the horizontal axis, measured counter-clockwise. Different curves correspond to different $\epsilon$'s. Notice that the optimal classifier changes as $\epsilon$ increases. We take $\sigma_1^2 = 2$ and $\sigma_2^2 = 0.2$.}
\label{FigAdvAccAngle}
 \end{figure}

\subsection{MNIST data}
We now describe the experiments on the MNIST dataset.
We use the variational bounds described in the Section~\ref{SecVariational} and implement the Fisher information term $\Phi(T|X)$ as a regularizer with coefficient $\beta$.
Recall that we considered $p_{T|X}(\cdot | x) = \cN(\mu(x; \theta), \Sigma(x; \theta))$, where $(\mu(x; \theta), \Sigma(x; \theta))$ are the mean and  (diagonal) covariance matrix of a $K$-dimensional Gaussian distribution. 
We evaluate the adversarial robustness of the neural networks using the Fast Gradient Sign Method (FGSM)~\cite{GooEtal14} with $\epsilon = 0.1$, with $10$ random initializations for each example.
As our model is inherently stochastic, we make the final prediction by taking an average over $12$ samples from the posterior.
This allows the adversary to obtain a consistent estimate of the gradient.  
We approximate the stochastic integral in equation~\eqref{EqnPhiVarGauss} with a single sample from the corresponding Gaussian distribution.
As both the loss function and the regularizer have a sum structure, we use the stochastic gradient-based Adam optimizer.
We implemented our experiments using Tensorflow~\cite{tensorflow2015} and the Adversarial robustness toolbox~\cite{art2018}.

\subsubsection{Single and multilayer neural networks}
\label{Sec:MnistRelu}

We first consider a simple one-layer architecture. The model architecture is $784-2K$ with $K = 10$, without any non-linearity.
 The first $K$ values of the last layer encode the mean, and the remaining $K$ values encode the variance of the features after a softplus transformation, similar to Alemi et al.~\cite{AleEtal16}.
Given the features $T=t$, the output of the classifier is a simple soft-max layer of the features (without any weights).
 The model is thus a variant of multiclass logistic regression with stochastic logits.
 For each value of $\beta$, we train the model for $150$ epochs.
Figure~\ref{FigMNISTLogistic} reports the effect of the regularization coefficient $\beta$ on clean accuracy and adversarial accuracy. Notice that as $\beta$ grows, the adversarial accuracy improves, while the test accuracy decreases. 
 
\begin{figure}[H]
    \centering
     \begin{minipage}{0.45\textwidth}
        \centering
		\scalebox{0.85}{\begin{tikzpicture}

\definecolor{color0}{rgb}{0.215686274509804,0.494117647058824,0.72156862745098}
\definecolor{color1}{rgb}{0.301960784313725,0.686274509803922,0.290196078431373}
\definecolor{color2}{rgb}{1,0.498039215686275,0}

\begin{axis}[
ticklabel style = {font=\tiny}, 
ytick distance = 0.1, 
legend style={cells={align=left}},
legend cell align={left},
legend style={at={(1,0.2)}, draw=white!80.0!black},
log basis x={10},
tick align=outside,
tick pos=left,
x grid style={white!69.01960784313725!black},
xlabel={$\displaystyle \beta$ (Regularization Coefficient)},
xmin=1.58489319246111e-08, xmax=6309573444.80194,
xmode=log,
xtick style={color=black},
y grid style={white!69.01960784313725!black},
ylabel={Test Accuracy},
ymin=0, ymax=1,
ytick style={color=black}
]
\addplot [ultra thick, color0, mark=diamond*, mark size=5, mark options={solid,fill=color1}]
table {%
1e-07 0.9149
1e-05 0.9148
0.0001 0.9116
0.001 0.8988
0.01 0.875
0.1 0.8681
1 0.8754
10 0.8624
100 0.8633
1000 0.8682
100000 0.8622
1000000 0.8632
10000000 0.8636
100000000 0.8576
1000000000 0.8577
};
\addlegendentry{Clean accuracy}
\addplot [very thick, color2, dashed]
table {%
1.58489319246111e-08 0.9139
6309573444.80199 0.9139
};
\addlegendentry{ $\beta = 0$ }
\end{axis}

\end{tikzpicture}}
        \caption*{(a) }
        \label{fig:prob1_6_1}
    \end{minipage}%
    \begin{minipage}{.55\textwidth}
        \centering
		\scalebox{0.85}{\begin{tikzpicture}

\definecolor{color0}{rgb}{0.215686274509804,0.494117647058824,0.72156862745098}
\definecolor{color1}{rgb}{0.301960784313725,0.686274509803922,0.290196078431373}
\definecolor{color2}{rgb}{1,0.498039215686275,0}

\begin{axis}[
ticklabel style = {font=\tiny}, 
legend style={cells={align=left}},
legend cell align={left},
legend style={at={(1,.2)}, draw=white!80.0!black},
log basis x={10},
tick align=outside,
tick pos=left,
x grid style={white!69.01960784313725!black},
xlabel={$\displaystyle \beta$ (Regularization Coefficient)},
xmin=1.58489319246111e-08, xmax=6309573444.80194,
xmode=log,
xtick style={color=black},
y grid style={white!69.01960784313725!black},
ylabel={Relative Adversarial Accuracy},
ymin=0.5, ymax=2.71806444525719,
ytick style={color=black}
]
\addplot [ultra thick, color0, mark=diamond*, mark size=5, mark options={solid,fill=color1}]
table {%
1e-07 1.0076869726181
1e-05 1.21872818470001
0.0001 1.59468913078308
0.001 2.20475196838379
0.01 2.46610760688782
0.1 2.52480792999268
1 2.49895191192627
10 2.53179621696472
100 2.54227805137634
1000 2.50943422317505
100000 2.54716992378235
1000000 2.54297709465027
10000000 2.53039836883545
100000000 2.59259271621704
1000000000 2.63661789894104
};
\addlegendentry{FGSM, $\epsilon = 0.1$}
\addplot[very thick, color2, dashed]
table {%
1.58489319246111e-08 1
6309573444.80199 1
};
\addlegendentry{ $\beta = 0$ }
\end{axis}

\end{tikzpicture}}
        \caption*{(b) }
        \label{fig:prob1_6_2}
    \end{minipage}%
\caption{
Plot showing the clean accuracy and  adversarial accuracy of the multi-class logistic regression as a function of the regularization coefficient $\beta$. Panel (a) reports the test accuracy of the model 
and panel (b) reports the relative adversarial accuracy as a function of $\beta$. We evaluate against the FGSM attack with $\epsilon = 0.1$ and $10$ random initializations. The baseline accuracy corresponds to the case when $\beta=0$, which is $14.31\%$.
}
    \label{FigMNISTLogistic}
\end{figure}

To show that this phenomenon is also observed in more complicated networks, 
we consider a simple fully-connected multilayer architecture: $784 -100 - 20 - 2K $. We use ReLU activations in all layers except the last layer, which is linear. 
Given the features, the output of the classifier is a simple soft-max layer of the features.

For each $\beta$, we train the network for $200$ epochs.
Figure~\ref{FigMnistReLu} shows the effect of $\beta$ on the adversarial accuracy.
The adversarial accuracy increases at first as we increase the regularization coefficient, supporting the claim that Fisher regularization leads to increased adversarial robustness.
If we further increase the regularization coefficient, increased robustness comes at the expense of the accuracy and leads to degraded performance.
 This trend is similar to the case of $\epsilon = 1.1$ in Figure~\ref{FigFisherAdvAcc}. We also tested this model against more powerful projected gradient descent (PGD) attacks. Although the absolute adversarial accuracy when using PGD attacks is lower compared to that obtained for an FGSM attack, the relative adversarial accuracy follows a trend identical to that in Figure~\ref{FigMnistReLu}.
\begin{figure}[H]
    \centering
     \begin{minipage}{0.45\textwidth}
        \centering
		\scalebox{0.85}{\begin{tikzpicture}

\definecolor{color0}{rgb}{0.215686274509804,0.494117647058824,0.72156862745098}
\definecolor{color1}{rgb}{0.301960784313725,0.686274509803922,0.290196078431373}
\definecolor{color2}{rgb}{1,0.498039215686275,0}

\begin{axis}[
ticklabel style = {font=\tiny}, 
ytick distance = 0.1, 
legend style={cells={align=left}},
legend cell align={left},
legend style={at={(1,0.3)}, draw=white!80.0!black},
log basis x={10},
tick align=outside,
tick pos=left,
x grid style={white!69.01960784313725!black},
xlabel={$\displaystyle \beta$ (Regularization Coefficient)},
xmin=4.46683592150963e-08, xmax=0.23872113856834,
xmode=log,
xtick style={color=black},
y grid style={white!69.01960784313725!black},
ylabel={Test Accuracy},
ymin=0.0, ymax=1.00722,
ytick style={color=black}
]
\addplot [ultra thick, color0, mark=diamond*, mark size=5, mark options={solid,fill=color1}]
table {%
1e-07 0.9643
1e-06 0.9649
1e-05 0.9594
0.0001 0.9489
0.001 0.9224
0.01 0.8319
0.1 0.5379
};
\addlegendentry{Clean accuracy}
\addplot [very thick, color2, dashed]
table {%
4.46683592150963e-08 0.9724
2.23872113856834 0.9724
};
\addlegendentry{ $\beta = 0$ }
\end{axis}

\end{tikzpicture}}
        \caption*{(a) }
        \label{fig:prob1_6_1}
    \end{minipage}%
    \begin{minipage}{.55\textwidth}
        \centering
		\scalebox{0.85}{\begin{tikzpicture}

\definecolor{color0}{rgb}{0.215686274509804,0.494117647058824,0.72156862745098}
\definecolor{color1}{rgb}{0.301960784313725,0.686274509803922,0.290196078431373}
\definecolor{color2}{rgb}{1,0.498039215686275,0}

\begin{axis}[
ticklabel style = {font=\tiny}, 
ytick distance = 1, 
legend style={cells={align=left}},
legend cell align={left},
legend style={at={(1,0.3)}, draw=white!80.0!black},
log basis x={10},
tick align=outside,
tick pos=left,
x grid style={white!69.01960784313725!black},
xlabel={$\displaystyle \beta$ (Regularization Coefficient)},
xmin=4.46683592150963e-08, xmax=0.23872113856834,
xmode=log,
xtick style={color=black},
y grid style={white!69.01960784313725!black},
ylabel={Relative Adversarial Accuracy},
ymin=0.5, ymax=10.7476913511753,
ytick style={color=black}
]
\addplot [ultra thick, color0, mark=diamond*, mark size=5, mark options={solid,fill=color1}]
table {%
1e-07 1.93461525440216
1e-06 2.44615387916565
1e-05 3.68846154212952
0.0001 6.94999980926514
0.001 8.73076915740967
0.01 10.2615375518799
0.1 6.17307710647583
};
\addlegendentry{FGSM, $\epsilon = 0.1$}
\addplot [very thick, color2, dashed]
table {%
4.46683592150963e-08 1
2.23872113856834 1
};
\addlegendentry{ $\beta = 0$ }
\end{axis}

\end{tikzpicture}}
        \caption*{(b) }
        \label{fig:prob1_6_2}
    \end{minipage}%
\caption{Plot showing the effect of $\beta$ on clean accuracy and adversarial accuracy. Panel (a) shows that clean accuracy decreases as we increase $\beta$. Panel (b) shows the relative  adversarial accuracy of the neural networks as a function of the regularization coefficient $\beta$. We evaluate adversarial accuracy against the FGSM attack with $\epsilon = 0.1$ and $10$ random initializations. The baseline accuracy corresponds to the case when $\beta=0$, which is $2.6\%$.
}
    \label{FigMnistReLu}
\end{figure}

\section{Discussion}
\label{SecDiscussion}

The research directions explored in this paper were inspired by recent work in adversarial machine learning. In particular, we were intrigued by the notion of a seemingly unavoidable tradeoff between robustness and accuracy, and the existence of a dichotomy between robust and non-robust features. A bottleneck formulation lends itself naturally to modeling a tradeoff between robustness and accuracy; measuring accuracy and robustness via information and estimation theoretic quantities, we have proposed the robust information bottleneck as a new variational principle for extracting maximally useful robust features. 

Like the standard information bottleneck, the robust information bottleneck formulation references only the data distribution, making it extremely general. Applying the principle for specific classes of features (e.g., linear or logistic) leads to feature-specific regularization terms. This means that one need not decide a priori to use an $\ell_1$- or $\ell_2$-regularizer, but instead use a regularization penalty corresponding to the Fisher information term discussed in this paper. Furthermore, we showed that the Fisher information term satisfies a host of properties that make it ideally suited to characterize robustness.

The robust information bottleneck is most clearly understood in the case of jointly Gaussian data. We showed that the optimally robust features in this setting is also jointly Gaussian with the data, and examined connections to the solution of the canonical information bottleneck.

From a practical point of view, we showed that it is computationally easy to extract features via the robust information bottleneck optimization using a variational approximation. We showed experimentally that a classifier trained on robust features extracted via the robust information bottleneck principle is indeed robust to simple adversarial attacks. Although we were able to defeat the classifier using stronger classes of adversarial attacks, our work in this paper suggests that a deeper investigation of Fisher regularization in neural networks is likely to be fruitful.

\section*{Acknowledgments}

This work was done in part while the authors were visiting the Simons Institute for the Theory of Computing during Summer 2019. VJ thanks Chandra Nair for pointing him to \cite[Corollary 1]{GengNair14}. AP acknowledges support from the UW-Madison Institute for Foundations of Data Science (IFDS), NSF grant CCF-1740707. VJ acknowledges support from NSF grants CCF-1841190 and CCF-1907786, the Wisconsin Alumni Research Foundation (WARF) through the Fall Research Competition Awards, and the Nvidia GPU grant program. PL acknowledges support from NSF grant DMS-1749857.

\bibliographystyle{unsrt}
\bibliography{ref.bib}

\appendix

\section{Proofs for Section~\ref{SecRIB}}

In this appendix, we provide the proofs of several lemmas introduced in the setup of the robust information bottleneck formulation.

\subsection{Proof of Lemma~\ref{lemma: invariance}}
\label{AppLemInvariance}

\begin{enumerate}
\item
Since $I(X; f(T)) = H(X) - H(X|f(T))$ and $I(X;T) =  H(X) - H(X|T)$, it is enough to show that $H(X|T) = H(X|f(T))$. This is a standard fact that may be found in Cover and Thomas~\cite{CovTho12}, but we include it here for completeness. If $S = f(T)$, then $p_S(s) ds = p_T(t)dt$ when $s = f(t)$. Hence,%
\begin{align*}
H(X|S) &= \int_{\real^d} H(X | S = s) p_S(s) ds\\
&= \int_{\real^d} H(X | S = f^{-1}(t)) p_T(t) dt\\
&= \int_{\real^d} H(X | T = t) p_T(t) dt\\
&= H(X|T).
\end{align*}
The proof for the equation involving $Y$ is analogous.

\item Note that $\mmse(Y|T)$ has the equivalent formulation 
\begin{align*}
\mmse(Y|T) = \inf_{g: \real^d \to \cY} \E[\| Y - g(T)) \|^2],
\end{align*}
where $g$ is any measurable function from $\real^d$ to $\cY$. For any such function $g$, we have
\begin{equation*}
\E[\| Y - g(T)) \|^2] = \E[\| Y - g(f^{-1}(S)) \|]^2.
\end{equation*}
Thus, taking an infimum over $g$, we obtain
\begin{align*}
\mmse(Y | T) = \inf_g \E[\| Y - g(f^{-1}(S)) \|^2 \geq \inf_g \E[\| Y - g(S)\|^2] = \mmse(Y | S).
\end{align*}
Analogously, we have $\mmse(Y|S) \geq \mmse(Y|T)$, proving the desired statement.

\item Observe that
\begin{align*}
\Phi(S|X) = \int \| \nabla_x \log p_{S|X}(s|x) \|_2^2 p_X(x) p_{S|X}(s|x) dx ds.
\end{align*}
Using the change of variables $t = f^{-1}(s)$, we have
\begin{align*}
\Phi(S|X) &= \int \left\| \nabla_x \log \frac{p_{T|X}(t|x)}{|\det \nabla_t f(t)|} \right\|_2^2 p_X(x) p_{T|X}(t|x) dx dt\\
&= \int \| \nabla_x \log p_{T|X}(t|x) \|^2 p_X(x) p_{T|X}(t|x) dx dt\\
&= \Phi(T|X),
\end{align*}
as wanted.
\item
Using the independence $T_2 \ind (T_1, X, Y)$, we have
\begin{equation*}
I(X; T) = I(X; T_1) + I(X; T_2|T_1) = I(X; T_1).
\end{equation*}
Similarly, we may obtain $I(Y; T) = I(Y; T_1)$. For the Fisher information term, observe that $$\nabla_x \log p_{T|X = x} (t) = \nabla_x (\log p_{T_1|X=x}(t_1) + \log p_{T_2}(t_2)) = \nabla_x \log p_{T_1|X=x}(t_1).$$ Taking the expectation of the squared-norms, we conclude that $\Phi(T|X) = \Phi(T_1|X)$. Finally, we have $p_{Y|T=t}(y) = p_{Y|T_1 = t_1}(y)$, so $\E[Y|T] = \E[Y|T_1]$. This shows that $\mmse(Y|T) = \mmse(Y|T_1)$.
\end{enumerate}

\subsection{Proof of Lemma~\ref{lemma: phi and mi}}
\label{AppLemPhiMI}

We use the entropic improvement of the Cram\'er-Rao bound (or Van Trees inequality) stated below:
\begin{equation}\label{EqnTom}
\frac{1}{2\pi e} \exp \left( \frac{2}{p} h(X|T) \right) \geq \frac{p}{J(X) + \Phi(T|X)}.
\end{equation}
The above inequality appeared recently in Aras et al.~\cite{AraEtal19}. Aras et al.~\cite{AraEtal19} noted that the entropic improvement does not appear to be well-known, although the univariate version was published  in Efroimovich~\cite{Efr80} in 1980. The claimed inequality immediately follows by noting that $I(X;T) = h(X) - h(X|T)$ and plugging in the lower bound for $h(X|T)$ from inequality~\eqref{EqnTom}.

\subsection{Proof of Lemma~\ref{LemFeatGauss}}
\label{AppLemFeatGauss}

Note that
\begin{equation*}
T \mid X = x \sim N(Ax, I),
\end{equation*}
so
\begin{equation*}
p_{T|X}(t|x) = \frac{1}{(2\pi)^{k/2}} \exp\left(-\frac{1}{2} (t-Ax)^\top  (t-Ax)\right).
\end{equation*}
In particular, we may calculate
\begin{equation*}
\nabla_x p_{T|X}(t|x) = \nabla_x\left(-\frac{1}{2}(t-Ax)^\top  (t-Ax)\right) = A^\top (t-Ax),
\end{equation*}
so
\begin{align*}
\Phi(T|X) & = \E_{T|X = x} \left[(t-Ax)^\top  AA^\top  (t-Ax)\right] = \E\left[\epsilon^\top  AA^\top  \epsilon\right] = \E_{\epsilon} \left[\tr(\epsilon^\top  AA^\top  \epsilon)\right] \\
& = \tr\left(AA^\top  \E[\epsilon \epsilon^\top ]\right) = \tr(AA^\top ) = \|A\|_F^2,
\end{align*}
as wanted.

\subsection{Proof of Lemma~\ref{LemFeatBin}}
\label{AppLemFeatBin}

We can compute
\begin{align*}
\nabla_x \mprob(T = 1 \mid X = x) & = - \nabla_x \log \left(1+\exp(-x^\top  w)\right) = \frac{\exp(-x^\top  w) w}{1+\exp(-x^\top w)}, \\
\nabla_x \mprob(T = -1 \mid X = x) & = - \nabla_x \log \left(1+\exp(x^\top  w)\right) = \frac{-\exp(x^\top  w) w}{1+\exp(x^\top w)},
\end{align*}
so that
\begin{align*}
\Phi(T|X = x) & = \frac{1}{1+\exp(-x^\top  w)} \cdot \frac{\exp(-2x^\top w) \|w\|_2^2}{\left(1+\exp(-x^\top w)\right)^2} + \frac{1}{1+\exp(x^\top w)} \cdot \frac{\exp(2x^\top w) \|w\|_2^2}{\left(1+\exp(x^\top w)\right)^2} \\
& = \|w\|_2^2 \left(\frac{\exp(-2x^\top w)}{\left(1+\exp(-x^\top w)\right)^3} + \frac{\exp(2x^\top w)}{\left(1+\exp(x^\top w)\right)^3}\right) \\
& = \|w\|_2^2 \cdot \frac{\exp(-2x^\top w) + \exp(-x^\top w)}{\left(1+\exp(-x^\top w)\right)^3} \\
& = \|w\|_2^2 \cdot \frac{\exp(-x^\top w)}{\left(1+\exp(-x^\top w)\right)^2} \\
& = \|w\|_2^2 \cdot \mprob(T = 1 \mid X = x) \cdot \mprob(T = -1 \mid X = x),
\end{align*}
as claimed.

\section{Proofs for Section~\ref{SecProperties}}
\label{AppProperties}

In this appendix, we provide proofs of the lemmas concerning robustness properties of Fisher information.

\subsection{Proof of Lemma~\ref{LemYhat}}
\label{AppLemYhat}

The proof is a direct application of the data processing inequality for Fisher information. We have the following sequence of inequalities:
\begin{align*}
\Phi(\widehat Y | X) &\stackrel{(a)}= J(X | \widehat Y) - J(X)\\
&\stackrel{(b)}\leq J(X | \widehat Y, T) - J(X)\\
&\stackrel{(c)}= J(X|T) - J(X)\\
&= \Phi(T|X).
\end{align*}
Here, $(a)$ follows from Lemma \ref{Lem:JPhi}, $(b)$ follows from Lemma \ref{Lem:JConvex}, and $(c)$ follows because $X \to T \to \widehat Y$ is a Markov chain.

\subsection{Proof of Lemma~\ref{LemGaussPerturb}}
\label{AppLemGaussPerturb}

Note that
\begin{align*}
I(X; T) - I(X+ \sqrt \delta Z; T) &= H(X)-H(X|T) - H(X+\sqrt \delta Z) + H(X+\sqrt \delta Z | T)\\
&= [H(X)- H(X+\sqrt \delta Z)] + [- H(X|T)  +[H(X+\sqrt \delta Z | T)]\\
& = -\frac{\delta}{2} J(X) + \frac{\delta}{2} J(X | T) + o(\delta) \\
&= \frac{\delta}{2} \Phi(T|X) + o(\delta),
\end{align*}
where the approximation follows by de Bruijn's identity~\cite{Sta59}.

\subsection{Proof of Lemma~\ref{LemKLPerturb}}
\label{AppLemKLPerturb}

Up to a second-degree approximation, we have
\begin{align*}
p_{T|X = x+ \epsilon u}(t) = p_{T|X = x}(t) + \epsilon (\nabla_x p_{T|X=x}(t) \cdot u)+ \epsilon^2( u^\top  \nabla^2_x p_{T|X=x} (t) u) + o(\epsilon^2).
\end{align*}
Thus, we may write
\begin{align}
\label{EqnKLPhi}
D(p_{T|X = x+ \epsilon u} \| p_{T|X = x}) &= \int p_{T|X = x+ \epsilon u}(t) \log \frac{p_{T|X = x+ \epsilon u}(t)}{p_{T|X = x}}dt \notag \\
&= \int \left[p_{T|X = x}(t) + \epsilon (\nabla_x p_{T|X=x}(t) \cdot u)+ \epsilon^2( u^\top  \nabla^2_x p_{T|X=x} (t) u) + o(\epsilon^2)\right] \times \notag \\
&\quad \quad \log \frac{p_{T|X = x}(t) + \epsilon (\nabla_x p_{T|X=x}(t) \cdot u)+ \epsilon^2( u^\top  \nabla^2_x p_{T|X=x} (t) u) + o(\epsilon^2)}{p_{T|X = x}}dt \notag \\
&= \int \left[p_{T|X = x}(t) + \epsilon (\nabla_x p_{T|X=x}(t) \cdot u)+ \epsilon^2( u^\top  \nabla^2_x p_{T|X=x} (t) u) + o(\epsilon^2)\right] \times \notag \\
&\quad \quad \log \left(1+  \frac{\epsilon (\nabla_x p_{T|X=x}(t) \cdot u)+ \epsilon^2( u^\top  \nabla^2_x p_{T|X=x} (t) u) + o(\epsilon^2)}{p_{T|X = x}}\right)dt \notag \\
&\stackrel{(a)}= \frac{\epsilon^2}{2} \left( \int p_{T|X = x}(t) \left(\frac{\nabla_x p_{T|X=x}(t) \cdot u}{p_{T|X = x}(t)}\right)^2 dt \right) + o(\epsilon^2) \notag \\
&\stackrel{(b)} = \frac{\epsilon^2}{2} \Phi(T | X = x) + o(\epsilon^2).
\end{align}
Here, in $(a)$ we have used the Taylor expansion of $\log(1+x) = x - \frac{x^2}{2} + o(x^2)$ and the equalities
\begin{align*}
\int u \cdot \nabla_x p_{T|X=x} (t) dt &= 0, \quad \text{and}\\
\int \left(u^\top  \nabla_x^2 p_{T|X=x}(t) u\right) dt &= 0,
\end{align*}
which hold under mild regularity conditions \cite[Lemma 5.3, pg. 116]{Leh06}. In $(b)$, we note that
\begin{align*}
 \int p_{T|X = x}(t) \left(\frac{\nabla_x p_{T|X=x}(t) \cdot u}{p_{T|X = x}(t)}\right)^2 dt &\leq  \int p_{T|X = x}(t) \left(\frac{\| \nabla_x p_{T|X=x}(t)\|}{p_{T|X = x}(t)}\right)^2 dt \\
 &= \Phi(T|X=x).
\end{align*}

\subsection{Proof of Lemma~\ref{LemRobBound}}
\label{AppLemRobBound}

Consider any $x, x' \in \mathcal{X}$ such that $\|x - x'\|_2 < \epsilon$. Note that for any $y$, we have
\begin{align*}
|\margin_f(x,y) - \margin_f(x',y)| \le 2 \left\|p_{T|X = x} (g(t) = \cdot) - p_{T|X = x'} (g(t) = \cdot) \right\|_{TV}.
\end{align*}
Indeed, we can write
\begin{align*}
& \margin_f(x,y) - \margin_f(x',y) \\
& \quad = \left(p_{T|X = x} (g(t) = y) - \max_{z \neq y} p_{T|X = x}(g(t) = z) \right) - \left(p_{T|X = x'} (g(t) = y) - \max_{z \neq y} p_{T|X = x'} (g(t) = z)\right) \\
& \qquad \le \left(p_{T|X = x} (g(t) = y) - p_{T|X = x} (g(t) = y')\right) - \left(p_{T|X = x'} (g(t) = y) - p_{T|X = x'} (g(t) = y')\right) \\
& \qquad \le \sum_{y} \left|p_{T|X = x} (g(t) = y) - p_{T|X = x'} (g(t) = y)\right| \\
& \qquad = 2 \left\|p_{Y|X = x} (g(t) = \cdot) - p_{Y|X = x'} (g(t) = \cdot) \right\|_{TV},
\end{align*}
where $y' = \arg\max_{z \neq y} p_{T|X = x'} (g(t) = z)$. A similar argument can be used to upper-bound the quantity $\margin_f(x',y) - \margin_f(x,y)$, from which the claim follows. In particular, Pinsker's inequality then implies that
\begin{align*}
2 \left\|p_{T|X = x} (g(t) = \cdot) - p_{T|X = x'} (g(t) = \cdot) \right\|_{TV} \le \sqrt{2 D\left( p_{T|X = x'} (g(t) = \cdot) \| p_{T|X = x} (g(t) = \cdot)\right)}.
\end{align*}
Furthermore, by the data processing inequality, we have
\begin{equation*}
D\left( p_{T|X = x'} (g(t) = \cdot) \| p_{T|X = x} (g(t) = \cdot)\right) \le D(p_{T|X = x'} \| p_{T|X = x} ).
\end{equation*}
Combining these inequalities with the bound~\eqref{EqnKLPhi}, we then have
\begin{align*}
\sup_{x' \in B_\epsilon(x)} |\margin_f(x,y) - \margin_f(x',y)| \le \sqrt{\epsilon^2 \Phi(T|X = x) + o(\epsilon^2)}.
\end{align*}
We can then integrate over $x$ to obtain
\begin{equation*}
\E\left[\sup_{x' \in B_\epsilon(x)} |\margin_f(x,y) - \margin_f(x',y)|^2 \right] \le \epsilon^2 \Phi(T|X) + o(\epsilon^2).
\end{equation*}

Define the event
\begin{equation*}
A^{\epsilon, \eta} := \left\{x \in \mathcal{X}: \sup_{x' \in B_\epsilon(x)} |\margin_f(x,f(x)) - \margin_f(x',f(x))|^2 \le \eta\right\}.
\end{equation*}
By Markov's inequality, we can write
\begin{equation*}
\mprob(x \notin A^{\epsilon, \eta}) = \mprob\left(\sup_{x' \in B_\epsilon(x)} |\margin_f(x,y) - \margin_f(x',y)|^2 > \eta\right) \le \frac{\epsilon^2 \Phi(T|X) + o(\epsilon^2)}{\eta}.
\end{equation*}
Note that if $x \in A^{\epsilon, \eta} \cap B^\delta$, we must have $\margin_f(x', f(x)) > 0$ for any $x' \in B_\epsilon(x)$, implying that $f(x) = f(x')$. Hence,
\begin{align*}
\mprob\left(f(x') = f(x) \quad \forall x' \in B_\epsilon(x)\right) & \ge \mprob(x \in A^{\epsilon, \eta} \cap B^\eta) \ge \mprob(x \in B^\eta) - \mprob(x \notin A^{\epsilon, \eta}) \\
& \ge \mprob(x \in B^\eta) - \frac{\epsilon^2 \Phi(T|X)+o(\epsilon^2)}{\eta},
\end{align*}
which is inequality~\eqref{EqnRobBound}.

\section{Proof of Gaussian optimality}\label{AppIBOptimality}

We now provide additional details on how to prove the optimality of Gaussian features when $X$ and $Y$ are jointly Gaussian, after having established a subadditivity result as in Lemmas~\ref{lemma: subaddinf} and~\ref{lemma: subaddmmse}. For brevity, we focus on the mutual information-based formulation of the robust information bottleneck; i.e., on Lemma \ref{lemma: subaddinf}. Throughout the proof, we shall use a common technique for smoothing densities to deal with technical issues. We do this by adding a small amount independent Gaussian noise to the random variables, and then taking the variance of the added Gaussian to 0 at the end~\cite{AnaEtal19}. %
Details of such proofs may be found in the references~\cite{GengNair14, AnaEtal19}. 

\begin{definition}
Let $X$ be an $\real^n$-valued random variable with density $p_X$. We say that $X$ (or equivalently, $p_X$) belongs the the set $\cP$ if:
\begin{enumerate}
\item
$\int_{\real^n} p_X(x) \log (1 + p_X(x)) dx < \infty$,
\item
$\E X = 0$, and
\item
$\E \|X\|^2 < \infty$.
\end{enumerate}
\end{definition}
All random variables in $\cP$ have well-defined entropies because of conditions (1) and (3). Condition (2) may be assumed without loss of generality, since the functions we consider are translation-invariant. Throughout the rest of the section, we shall assume that that the random variables under consideration lie in $\cP$.

Consider the function $f_\delta$ defined on the space of densities in $\cP$ as
\begin{align*}
f_\delta(X) &= -H(C(X + \sqrt \delta N)+\xi) + \gamma H(X+\sqrt \delta N) - \beta J(X + \sqrt \delta N)\\
&:= -H(Y_\delta) + \gamma H(X_\delta) - \beta J(X_\delta),
\end{align*}
where $N \ind X$ and $N \sim \cN(0, I)$. The upper-concave envelope of $f_\delta$ is given by
\begin{align*}
F_\delta(X) = \sup_{p_{U|X}(\cdot | \cdot)} \left\{- H(Y_\delta|T) + \gamma H(X_\delta|T) - \beta J(X_\delta|T)\right\}.
\end{align*}
We define a lifting of $f$ to pairs of random variables, as follows:
\begin{align*}
f_\delta(X_1, X_2) = -H(Y_{1\delta}, Y_{2\delta}) + \gamma H(X_{1\delta}, X_{2\delta}) - \beta J(X_{1\delta}, X_{2\delta}),
\end{align*}
where we use $X_{i\delta} = X_i + \sqrt \delta N_i$, and $Y_{i\delta} = CX_{i\delta} + \xi_i$ for $i \in \{1, 2\}$. Note that $(N_1, N_2, \xi_1, \xi_2)$ are independent of $(X_1, X_2, U)$ and also of each other. Let $F_\delta(X_1, X_2)$ be the upper-concave envelope of $f_\delta(X_1, X_2)$. Just as in Lemma~\ref{lemma: subaddinf}, we are able to show a subadditivity result for $F_\delta(\cdot, \cdot)$:

\begin{lemma*}
\label{lemma: subaddinf_appendix}
For any pair of random variables $(X_1, X_2)$, we have that 
\begin{align*}
F_\delta(X_1, X_2) \leq F_\delta(X_1) + F_\delta(X_2).
\end{align*}
\end{lemma*}

\begin{proof}
Note that $N_1$ and $N_2$ are independent of $(X_1, X_2, T)$ and also of each other. The proof is identical to that of Lemma \ref{lemma: subaddinf}, so we omit the details here.
\end{proof}

We also have a simple corollary of Lemma \ref{lemma: subaddinf_appendix}:

\begin{corollary*}
If $X_1$ and $X_2$ are independent random variables, then 
$$F_\delta(X_1, X_2) = F_\delta(X_1) + F_\delta(X_2).$$
\end{corollary*}

\begin{proof}
Using Lemma \ref{lemma: subaddinf}, we have 
\begin{align*}
F_\delta(X_1, X_2) \leq F_\delta(X_1) + F_\delta(X_2).
\end{align*}
To prove the reverse direction, notice that we may always choose $U = (U_1, U_2)$ such that $$(X_1, U_1) \ind (X_2, U_2).$$ 
For such a $U$, we have
\begin{align*}
f_\delta(X_1, X_2 | U) = f_\delta(X_1 | U_1) + f_\delta(X_2 | U_2).
\end{align*}
Taking the supremum of the right hand side over all $U_1$ and $U_2$, we obtain 
\begin{align*}
F_\delta(X_1, X_2) \geq \sup_{(U_1, U_2)} f_\delta(X_1, X_2 | U) = F_\delta(X_1) + F_\delta(X_2),
\end{align*}
which proves the claim.
\end{proof}

More importantly, we can also prove a ``converse'' to the above corollary:

\begin{lemma*}
\label{lemma: independence}
If $(X_1, X_2)$ and $U$ are such that
\begin{enumerate}
\item[(a)]
$F_\delta(X_1, X_2) = F_\delta(X_1) + F_\delta(X_2)$, and 
\item[(b)]
$F_\delta(X_1, X_2) = f_\delta(X_1, X_2 | U)$,
\end{enumerate}
then the following results hold:
\begin{enumerate}
\item
For all $u \in \supp(U)$, we have $X_1 \ind X_2$ conditioned on $U=u$, and
\item
$f_\delta(X_1 | U) = F_\delta(X_1)$ and $f_\delta(X_2 | U) = F_\delta(X_2).$
\end{enumerate}
\end{lemma*}

\begin{proof}
The key point is to identify the conditions under which equality holds in the proof of Lemma \ref{lemma: subaddinf}. We see that equality holds if and only if 
\begin{align*}
J(X_{1\delta} | X_{2\delta}, U) &= J(X_{1\delta} | U), \quad \text{and}\\
H(Y_{2\delta}|Y_{1\delta}, U) &= H(Y_{2\delta}|X_{1\delta}, U).
\end{align*}
The convexity of $J(\cdot)$ from Lemma~\ref{Lem:JConvex} implies that $X_{1\delta} \ind X_{2\delta}$ conditioned on all values of $U = u \in \supp(U)$, and thus proves (1). It is not hard to show (using characteristic functions, for example) that $X_{1\delta} \ind X_{2\delta}$ implies $X_1 \ind X_2$. Now observe that
\begin{align*}
f_\delta(X_1, X_2 | U) &= f_\delta(X_1 | U) + f_\delta(X_2 | U).
\end{align*}
The above equality, combined with the assumed equality $f_\delta(X_1, X_2 | U) = F_\delta(X_1, X_2)$, immediately yields the equalities in $(2)$.
\end{proof}

The rest of the proof closely follows the steps outlined in Geng and Nair~\cite[Appendix II]{GengNair14} and Anantharam et al.~\cite[Section 4.3]{AnaEtal19}. 

\begin{definition}\label{def: vV}
Define
\begin{align}
v (\Sigma) &:= \sup_{\Cov(X) = \Sigma} f_\delta(X), \quad \text{ and } \label{eq: vV-v}\\
V(\Sigma) &:= \sup_{\Cov(X) \preceq \Sigma} F_\delta(X). \label{eq: vV-V}
\end{align}
\end{definition}

\begin{lemma*}
\label{lemma: xstar_ustar}
There exist random variables $X^*$ and $U^*$ satisfying
\begin{enumerate}
\item $| \cU^*| \leq  \frac{n(n+1)}{2} +1$, and
\item $\Cov(X^*) \preceq \Sigma$,
\end{enumerate}
such that the following holds:
\begin{equation}
V(\Sigma) = f_\delta(X^* | U^*).
\end{equation}
\end{lemma*}

\begin{proof}[Proof sketch]
Let $(X^{(t)}, t \ge 1)$ be a sequence of random variables such that $\Cov X^{(t)} = \widehat \Sigma$ and $f_\delta(X^{(t)}) \uparrow v (\widehat \Sigma)$ as $t \to \infty$. Using tightness of this sequence \cite[Proposition 17]{GengNair14}, we set $X^{\widehat \Sigma}$ to be the weak limit of $X^{(t)}$ as $t \to \infty$. Since $X^{(t)}_\delta$ satisfies the necessary regularity conditions as in Geng and Nair~\cite[Proposition 18]{GengNair14}, we  have $h(X^{(t)}_\delta) \to h(X^{\widehat \Sigma}_\delta)$, $h(Y^{(t)}_\delta) \to h(Y^{\widehat \Sigma}_\delta)$. Moreover, we also have $J(X^{(t)}_\delta) \to J(X^{\widehat \Sigma}_\delta)$ (see Lemma \ref{lemma: fisher limit}). Hence, we may conclude that $f_\delta(X^{\widehat \Sigma}) = v(\widehat \Sigma)$.

Recall that $V(\Sigma)$ is defined as
\begin{align}
V(\Sigma) &= \sup_{\Cov(X) \preceq \Sigma} F_\delta(X) \nonumber \\
&= \sup_{(U, X), \Cov(X) \preceq \Sigma} f_\delta(X \mid U) \nonumber \\
&\stackrel{(a)}= \sup_{\alpha_l \geq 0, \widehat \Sigma_l : \sum_{l=1}^M \alpha_l = 1, \sum_{l=1}^M \alpha_l \widehat \Sigma_l \preceq \Sigma} \sum_{l=1}^M \alpha_l v(\widehat \Sigma_l), \label{eq: Caratheodory}
\end{align}
where, for the moment, $M$
ranges over positive integers of
arbitrary size. The equality in $(a)$ is because we may restrict $p_{X|T}(\cdot|U)$ to the class of optimizers $X^{\widehat \Sigma}$ for $\widehat \Sigma \succeq 0$.  Fenchel's extension of Carath\'{e}odory's Theorem \cite[Theorem 1.3.7]{HUT} implies that we can fix $M$ to be $ {{n+1} \choose 2}+1$
in equation~\eqref{eq: Caratheodory} (see Lemma 4.3 in Anantharam et al.~\cite{AnaEtal19} for details.)

Consider any sequence of convex combinations $\left( \{\alpha_l^{(t)}\}_{l=1}^M, \{\widehat \Sigma_l^{(t)}\}_{l=1}^M \right)$ 
with $\sum_{l=1}^m \alpha_l^{(t)} \widehat \Sigma_l^{(t)} \preceq \Sigma$ for all $t \ge 1$, and such that $\sum_{l=1}^m \alpha_l^{(t)} v(\widehat \Sigma_l^{(t)})$ converges to $v(\Sigma)$ as
$t \to \infty$. Since the $M$-dimensional simplex is compact, we may assume that $\alpha_l^{(t)} \to \alpha^*_l$ for all $l \in [M]$. Suppose that for some $l$, we have $\alpha_l^{(t)} \to 0$ as $t \to \infty$. Then noticing that $\alpha_l^{(t)} \widehat \Sigma_l^{(t)} \preceq \Sigma$, we have
\begin{align*}
v(\widehat \Sigma_l^{(t)}) &\stackrel{(a)}\leq   -h(\xi) + \frac{\gamma}{2} \log (2\pi e)^n \det (\widehat \Sigma^{(t)}_l + \delta I) - \beta \tr((\widehat \Sigma^{(t)}_l + \delta I)^{-1})\\
&\stackrel{(b)}\leq -h(\xi) + \frac{\gamma}{2} \log (2\pi e)^n \det ( \frac{\Sigma}{\alpha^{(t)}_l} + \delta I) - \beta \tr((\frac{\Sigma}{\alpha^{(t)}_l} + \delta I)^{-1})\\
&\stackrel{(c)}= -h(\xi) + \frac{\gamma}{2} \log (2\pi e)^n \prod_{i=1}^n \left( \frac{\lambda_i}{\alpha^{(t)}_l} + \delta \right) - \beta \sum_{i=1}^n \left(\frac{\lambda_i}{\alpha^{(t)}_l} + \delta\right)^{-1}.
\end{align*}
In $(a)$, we used
$h(Y_\delta) \geq h(\xi)$, and the fact that Gaussians maximize entropy and minimize Fisher information for a fixed variance. In $(b)$, we used $\alpha_l^{(t)} \widehat \Sigma_l^{(t)} \preceq \Sigma$, and in $(c)$ we let $\lambda_i$ for $i \in [n]$ be the eigenvalues of $\Sigma$. It is now clear that if $\alpha^{(t)}_l \to 0$, then $\alpha^{(t)}_l  v(\widehat \Sigma^{(t)}_l) \to 0$, as well. So we may assume without loss of generality that $\min_{l \in [M]} \alpha^*_l = \alpha_{\min} > 0$. This implies that $\widehat \Sigma^{(t)}_l \preceq \frac{2\Sigma}{\alpha_{\min}}$ for all large enough $t$, so we can find a convergent subsequence such that $\widehat \Sigma^{(t)}_l \rt \Sigma_l^*$ for each $1 \le l \le M$ when $t \to \infty$. This leads us to the equation
\begin{equation*}
V(\Sigma) = \sum_{l=1}^M \alpha_l^* v(\Sigma^*_l).
\end{equation*}
In other words, we can find a pair of random variables $(X^*, U^*)$ with $|\cU^*| \leq M$ such that $V(\Sigma) = f_\delta(X^* | U^*)$. This completes the proof.
\end{proof}
Having shown existence, we note a simple property stated below that is crucial for establishing Gaussian optimality:

\begin{lemma*}
\label{lemma: rotation}
Consider random variables $(X_1,X_2, U)$, and define new random variables $X_+$ and $X_-$ via
\begin{align*}
X_+ := \frac{X_1 + X_2}{\sqrt 2}, \quad &\text{ and } \quad X_- := \frac{X_1 - X_2}{\sqrt 2}.
\end{align*}
Then $f_\delta(X_1, X_2 | U) = f_\delta(X_+, X_-|U)$.
\end{lemma*}

\begin{proof}[Proof sketch]
The proof follows by noticing that entropy and Fisher information functionals are invariant to unitary transformations, and we omit the details here.
\end{proof}
In the next lemma, we use all of the prior lemmas in a remarkable sequence of inequalities that yield Gaussian optimality.

\begin{lemma*}
\label{lemma: gaussian_delta}
Let the random variables $X^*$ and $U^*$ be as in Lemma \ref{lemma: xstar_ustar}; i.e., satisfying the equality $V(\Sigma) = f_\delta(X^* | U^*)$, and with $|\cU^*| \leq M$. Consider two independent and identically distributed copies of $(X^*,U^*)$, denoted by $(X_1, U_1)$ and $(X_2, U_2)$. Define new random variables $X_+$ and $X_-$, as follows:
\begin{align*}
X_+ := \frac{X_1 + X_2}{\sqrt 2}, \quad &\text{ and } \quad X_- := \frac{X_1 - X_2}{\sqrt 2}.
\end{align*}
Also define $U \defn (U_1, U_2)$. Then the following results hold:
\begin{enumerate}
    \item [(a)] $X_+$ and $X_-$ are conditionally independent given $U$.
    \item [(b)] $V(\Sigma) = f_\delta(X_+|U)$ and $V(\Sigma) = f_\delta(X_-|U)$.
\end{enumerate}
\end{lemma*}

\begin{proof}
We have the following sequence of inequalities:
\begin{align*}
    2V(\Sigma) &\stackrel{(a)}= f_\delta(X_1 | U_1) + f_\delta(X_2 | U_2)\\
    &\stackrel{(b)}= f_\delta(X_1, X_2 | U_1, U_2)\\
    &\stackrel{(c)}= f_\delta(X_+, X_- | U_1, U_2)\\
    &\stackrel{(d)}\leq F_\delta(X_+, X_-)\\
    &\stackrel{(e)}\leq F_\delta(X_+) + F_\delta(X_-)\\
    &\stackrel{(f)}\leq V(\Sigma) + V(\Sigma) = 2V(\Sigma).
\end{align*}
Here, $(a)$ follows from the assumption that $f_\delta(X^*| U^*) = V(\Sigma)$. Equality $(b)$ follows from the independence $(X_1, U_1) \ind (X_2, U_2)$. Equality $(c)$ holds because of Lemma \ref{lemma: rotation}. Inequality $(d)$ follows from the definition of $F_\delta(\cdot)$. Inequality $(e)$ follows from the Lemma \ref{lemma: subaddinf_appendix}. Finally, inequality $(f)$ follows from the definition in equation \eqref{eq: vV-V}, and the fact that $X_+$ and $X_-$ have the same covariance as $X^*$, which is bounded above by $\Sigma$ in the positive semidefinite partial order.

Since the first and last expressions match, all the inequalities in the above sequence of inequalities must be equalities. In particular, equalities $(d)$ and $(e)$ combined with Lemma \ref{lemma: independence} imply that $X_+ \ind X_-$ conditioned on $(U_1, U_2)$, thus establishing part  (a) of the lemma. Lemma \ref{lemma: independence} also gives $f_\delta(X_+|U_1, U_2) = F_\delta(X_+)$ and $f_\delta(X_-|U_1, U_2) = F_\delta(X_-)$. Finally, equality in $(f)$ gives $F_\delta(X_+) = V(\Sigma)$ and $F_\delta(X_-) = V(\Sigma)$. This completes the proof of part (b).
\end{proof}
Finally, we show that the optimizer may be chosen to be a single and unique Gaussian distribution.

\begin{lemma*}
\label{lemma: single_gaussian}
There exists $G^* \sim \cN(0, \Sigma^*) $ such that $\Sigma^* \preceq \Sigma$ and $V(\Sigma) = f(G^*)$. Furthermore, the random variable $G^*$ is the unique  zero-mean element with covariance less than $\Sigma$ satisfying $f_\delta(X) = V(\Sigma)$.
\end{lemma*}
\begin{proof}
The proof is identical in all respects to that of Geng and Nair \cite[Theorem 1]{GengNair14} and Anantharam et al.~\cite[Lemma 4.6]{AnaEtal19}, so we omit it here.
\end{proof}

\begin{lemma*}
\label{lemma: fisher limit}
Consider a sequence of random variables $\{X^{(t)}\}_{t \geq 0}$ such that $X^{(t)}$ converges weakly to $X$ as $t \to \infty$. As before, denote $X^{(t)} + \sqrt \delta N := X^{(t)}_\delta$, and similarly for $X_\delta$. Then we have
\begin{align*}
\lim_{t \to \infty} J(X^{(t)}_\delta) = J(X_\delta).
\end{align*}
\end{lemma*}

\begin{proof}
Let $G$ be a standard normal random variable that is independent of $X$. For $\tau \in (0, \infty)$, the function $h(X+\sqrt \tau G)$ is known to be a smooth, monotonically increasing, concave function of $\tau$ \cite{Cos85, Vil00}. Furthermore, de Bruijn's identity \cite{Bla65} states that
\begin{align*}
\frac{d}{d\tau} h(X + \tau G) \Big|_{\tau = \tau_0} = \frac{1}{2} J(X+ \sqrt \tau_0 G).
\end{align*}
Using the concavity and monotonicity properties of this function, we observe that for any $t \geq 0$, any $\eta > 0$, and any $\delta' \in (0, \delta)$, we have
\begin{align*}
\frac{1}{2} J(X^{(t)}_\delta) &\geq \frac{h(X^{(t)}_\delta + \sqrt \eta G) - h(X^{(t)}_\delta)}{\eta}, \quad \text{ and }\\
\frac{1}{2} J(X^{(t)}_\delta) &\leq \frac{h(X^{(t)}_\delta) - h(X^{(t)}_{\delta'})}{\delta-\delta'},
\end{align*}
where $\delta' \in (0, \delta)$. Taking the $\lim\inf$ in the first inequality and $\lim\sup$ in the second as $t \to \infty$, we obtain
\begin{align*}
\lim\inf_{t \to \infty} \frac{1}{2} J(X^{(t)}_\delta) &\geq \frac{h(X_\delta + \sqrt \eta G) - h(X_\delta)}{\eta}, \quad \text{ and }\\
\lim\sup_{t \to \infty}\frac{1}{2} J(X^{(t)}_\delta) &\leq \frac{h(X_\delta) - h(X_{\delta'})}{\delta-\delta'}.
\end{align*}
Since the above statements are valid for any choice of $\eta>0$ and $\delta' \in (0, \delta)$, we may take the limit as $\eta \to 0$ and $\delta' \to \delta$ to conclude that
\begin{align*}
\lim_{t \to \infty}\frac{1}{2} J(X^{(t)}_\delta) = \frac{1}{2} J(X_\delta).
\end{align*}
\end{proof}

Finally, we shall take $\delta \to 0$ and establish Gaussian optimality results for the function $f(\cdot)$.

\begin{lemma*}
Define the function $f:\cP \to \real$ as
\begin{align}
f(X) = -H(CX+\xi) + \gamma H(X) - \beta J(X) := -H(Y) + \gamma H(X) - \beta J(X),
\end{align}
and consider the optimization problem 
\begin{align*}
\sup_{\Cov(X) \preceq \Sigma}f(X).
\end{align*}
The supremum of the above optimization problem is attained by a Gaussian random variable with covariance $\Sigma^* \preceq \Sigma$.
\end{lemma*}

\begin{proof}
Let $X$ be such that $\Cov(X) \preceq K$. We know that
\begin{align*}
f_\delta(X) &\leq \sup_{G \sim \cN(0, \Sigma), \Sigma \preceq K} f_\delta(G) \\
&=\sup_{\Sigma \preceq K} -\frac{1}{2} \log (2\pi e)^k |C(\Sigma + \delta I) C^\top  + \Sigma_\xi| + \frac{\gamma}{2} \log (2 \pi e)^n |\Sigma + \delta I| - \frac{\beta}{2} \tr (\Sigma + \delta I)^{-1}.
\end{align*}
As $\delta \to 0$, we have 
\begin{align*}
\lim_{\delta \to 0} h(X_\delta) &= h(X),\\
\lim_{\delta \to 0} h(CX_\delta + \xi) &= h(CX + \xi),\\
\lim\sup_{\delta} J(X_\delta) &\leq J(X).
\end{align*}
The first two equalities follow from the bounded variance assumption on $X$, and the last inequality follows from the convexity of $J(\cdot)$. We cannot assert the equality of the Fisher information limit without making additional smoothness assumptions on the density of $X$; however, inequality is sufficient for our purposes here, since we may conclude that
\begin{align*}
f(X) \leq \lim\inf_{\delta \to 0} f_\delta(X).
\end{align*}
However, we also have
\begin{multline*}
\lim\inf_{\delta \to 0} f_\delta(X) 
\leq \lim\inf_{\delta \to 0} \sup_{\Sigma \preceq K} -\frac{1}{2} \log (2\pi e)^k |C(\Sigma + \delta I) C^\top  + \Sigma_\xi| \\
+ \frac{\gamma}{2} \log (2 \pi e)^n |\Sigma + \delta I| - \frac{\beta}{2} \tr (\Sigma + \delta I)^{-1}.
\end{multline*}
We claim that as $\delta \to 0$, the following equality holds:
\begin{align*}
\lim_{\delta \to 0}&\Bigg[ \sup_{\Sigma \preceq K} -\frac{1}{2} \log (2\pi e)^k |C(\Sigma + \delta I) C^\top  + \Sigma_\xi| + \frac{\gamma}{2} \log (2 \pi e)^n |\Sigma + \delta I| - \frac{\beta}{2} \tr (\Sigma + \delta I)^{-1}\Bigg]\\
&= \Bigg[ \sup_{\Sigma \preceq K} -\frac{1}{2} \log (2\pi e)^k |C\Sigma C^\top  + \Sigma_\xi| + \frac{\gamma}{2} \log (2 \pi e)^n |\Sigma + \delta I| - \frac{\beta}{2} \tr \Sigma^{-1}\Bigg] \\
& = \sup_{G \sim \cN(0, \Sigma), \Sigma \preceq K} f(G).
\end{align*}
To simplify notation, denote 
\begin{align*}
\theta_\delta(\Sigma) &= \Bigg[ -\frac{1}{2} \log |C(\Sigma + \delta I) C^\top  + \Sigma_\xi| + \frac{\gamma}{2} \log |\Sigma + \delta I| - \frac{\beta}{2} \tr (\Sigma + \delta I)^{-1}\Bigg], \\
\theta(\Sigma) &= \Bigg[ -\frac{1}{2} \log  |C\Sigma C^\top  + \Sigma_\xi| + \frac{\gamma}{2} \log |\Sigma | - \frac{\beta}{2} \tr (\Sigma^{-1})\Bigg].
\end{align*}
Notice that for any fixed $\Sigma \succeq 0$, we have $\Theta_\delta(\Sigma) \to \Theta(\Sigma)$; i.e., we have pointwise convergence of these two functions on $\left\{\Sigma: \Sigma \preceq K\right\}$. Let $M = \Theta(K)-\epsilon_0$ for some small $\epsilon_0 > 0$ and pick $\delta_0$ small enough so that $\Theta_\delta(K) \geq M$ for all $\delta < \delta_0$. 

Notice that we are concerned with the maxima of $\Theta(\cdot)$ and $\Theta_\delta(\cdot)$ on the space $\Sigma \preceq K$. Without loss of generality, we may restrict to the space 
$$\cR := \left\{\Sigma ~:~ \Sigma \preceq K \right\} \cap \left\{ \Sigma: \lambda_{\min}(\Sigma) > \lambda_{\min}^* - \delta_0 \right\},$$ 
for a careful choice of $\lambda_{\min}^*$ and any small enough $\delta_0 < \lambda_{\min}^*$. This is because we have
\begin{align*}
\Theta(\Sigma) &\leq  -\frac{1}{2} \log  |\Sigma_\xi| + \frac{\gamma}{2} \log |\Sigma| - \frac{\beta}{2} \tr (\Sigma^{-1})\\
&= -\frac{1}{2} \log  |\Sigma_\xi| + \frac{\gamma}{2} \sum_{i=1}^n \log \lambda_i - \beta \sum_{i=1}^n \frac{1}{\lambda_i}.
\end{align*}
Note that $\lambda_{\max}(\Sigma) \leq \lambda_{\max} (K)$, so if $\lambda_i \to 0$ for any $i$, we will have $\Theta(\Sigma) \to -\infty$. So by imposing some threshold ($M$ in this instance), we may rule out any matrix $\Sigma$ whose smallest eigenvalue is so small that $\Theta(\Sigma) < M$, hence is not a contender for the supremum of $\Theta(\cdot)$. Suppose the lower bound obtained in this way is $\lambda_{\min}^*$. Observing that $\Theta_\delta(\Sigma) = \Theta(\Sigma + \delta I)$, we see that the eigenvalues of $\Sigma+\delta I$ should also be larger than $\lambda_{\min}^*$, so we may define the region $\cR$ as shown above.

 It is now easy to check that over the set $\cR$, the function $\Theta(\cdot)$ and $\Theta_\delta(\cdot)$ converge uniformly, implying that
\begin{align*}
\sup_{\Sigma \preceq K}  \Theta(\Sigma) &= \sup_{\Sigma \in \cR}  \Theta(\Sigma)
= \lim_{\delta \to 0} \sup_{\Sigma \in \cR} \Theta_\delta(\Sigma) = \lim_{\delta \to 0} \sup_{\Sigma \preceq K} \Theta_\delta(\Sigma),
\end{align*}
which proves the claim.
\end{proof}

\section{Derivations for Gaussians}

In this appendix, we provide the details of the linear-algebraic derivations used to obtain the formulas for the optimal feature maps in the theorems of Section~\ref{SecGaussian}.

\subsection{Proof of Theorem~\ref{ThmInfoId}}
\label{AppThmInfoId}

Let $(X_G,Y_G)$ be jointly Gaussian with $\Cov(X_G) = K$ and $Y_G = CX_G + \xi$, where $\Cov(\xi)= \Sigma_\xi = \Sigma_y - \Sigma_{yx} \Sigma_x^{-1} \Sigma_{xy}$. From Lemma~\ref{lemma: subaddinf} in the proof of Gaussian optimality, we see that it is enough to solve the optimization problem
\begin{align*}
\sup_{X \sim \cN(0, \Sigma), \Sigma \preceq K} \left\{- H(CX+\xi) + \gamma H(X) - \beta J(X)\right\},
\end{align*}
since the optimum $X^*$ of the above expression can be used to deduce the optimum $T$ for $(X_G, Y_G)$. For an $\real^d$-valued Gaussian random variable with covariance $\Sigma$, we have closed-form expressions for the entropy and Fisher information: $H(X) = \frac{d}{2} \log 2\pi e + \frac{1}{2} \log \det \Sigma$ and $J(X) = \tr(\Sigma^{-1}).$ Using these formulas and ignoring constant terms, we can rewrite the optimization problem as
\begin{align*}
\inf_{\Sigma \preceq K} \left\{\frac{1}{2} \log \det(C\Sigma C^\top  + \Sigma_\xi) - \gamma \frac{1}{2} \log \det (\Sigma) + \beta \tr(\Sigma^{-1})\right\}.
\end{align*}
Setting $B:= \Sigma_\xi^{-1/2} C$ and ignoring terms that do not depend on $\Sigma$, we further obtain
\begin{align*}
\inf_{\Sigma \preceq K} \left\{\frac{1}{2} \log \det(B\Sigma B^\top  + I) - \gamma \frac{1}{2} \log \det (\Sigma) + \beta \tr(\Sigma^{-1})\right\}.
\end{align*}
We now plug in $K = \sigma_x^2 I$. Writing the SVD $\frac{\Sigma}{\sigma_x^2} = UDU^\top $, we have the equivalent optimization problem
\begin{equation}
\label{EqnObjSep}
\min_{\stackrel{0 \preceq D \preceq I}{U^\top U = I = UU^\top }} \left\{\frac{1}{2} \log \det(\sigma_x^2 BUDU^\top  B^\top  + I) - \frac{\gamma}{2} \log \det(D) + \frac{\beta}{\sigma_x^2} \tr(D^{-1})\right\}.
\end{equation}

We now write the SVD $B = V \Lambda W^\top $, where $V \in \real^{k \times k}$ is an orthogonal matrix and $W \in \real^{p \times k}$ satisfies $W^\top W = I_k$. Then
\begin{equation*}
\det(\sigma_x^2 BUDU^\top  B^\top  + I) = \det\left(V(\sigma_x^2 \Lambda W^\top  U D U^\top  W \Lambda + I)V^\top \right) = \det(\sigma_x^2 \Lambda W^\top  U D U^\top  W \Lambda + I).
\end{equation*}
Note in particular that this is the only term in the expression~\eqref{EqnObjSep} which depends on $U$, so we fix $D$ for the time being and focus on optimizing this term with respect to $U$. Let $\tilde{U} = U^\top W$. Note that $\tilde{U}^\top  \tilde{U} = W^\top UU^\top W = W^\top W = I$, and furthermore,
\begin{equation*}
\det(\sigma_x^2 \Lambda \tilde{U}^\top  D \tilde{U} \Lambda + I) = (\det(\Lambda))^2 \det(\sigma_x^2 \tilde{U}^\top  D \tilde{U} + \Lambda^{-2}).
\end{equation*}
We will apply Lemma~\ref{LemFan} with $A = \sigma_x^2 \tilde{U}^\top D\tilde{U}$ and $B = \Lambda^{-2}$. Let
\begin{align*}
(a_1, \dots, a_k) & := \lambda(A)^{\downarrow}, \\
(\ell_1, \dots, \ell_k) & := \lambda(B)^{\downarrow}.
\end{align*}
We first have the following lemma:

\begin{lemma*}
\label{LemMaj}
Let $(d_1, \dots, d_p) := \lambda(D)^{\downarrow}$. We have $\lambda(A)^{\downarrow} \preceq \sigma_x^2 \cdot (d_1, \dots, d_k)$.
\end{lemma*}

\begin{proof}
For $1 \le j \le k$, we can write
\begin{equation*}
\sum_{i=1}^j a_i = \tr(V^\top  A V),
\end{equation*}
where the columns $V \in \real^{k \times j}$ are orthonormal eigenvectors of $A$ with eigenvalues $\{a_1, \dots, a_j\}$. Furthermore,
\begin{equation*}
\tr(V^\top AV) = \tr(\sigma_x^2 V^\top  \tilde{U}^\top  D \tilde{U} V) \le \max_{U \in \real^{k \times j}: U^\top U = I_j} \tr(\sigma_x^2 U^\top DU),
\end{equation*}
since $(\tilde{U} V)^\top  (\tilde{U} V) = V^\top  \tilde{U}^\top  \tilde{U} V = V^\top V = I$. Finally, note that Lemma~\ref{LemMaxEigs} implies that the last quantity is equal to $\sigma_x^2 \cdot \sum_{i=1}^j d_i$. Thus, the desired result follows.
\end{proof}

Thus, Lemmas~\ref{LemMaj} and~\ref{LemFan} together imply that
\begin{equation}
\label{EqnMaj}
\lambda\left(\sigma_x^2 \tilde{U}^\top  D \tilde{U} + \Lambda^{-2}\right)^{\downarrow} \preceq \lambda(A)^{\downarrow} + \lambda(B)^{\downarrow} \preceq \sigma_x^2 \cdot (d_1, \dots, d_k) + \lambda(B)^{\downarrow}.
\end{equation}
Furthermore, recall that the function $f: \real_+^p \rightarrow \real$ defined by $f(x) = \prod_{i=1}^p x_i$ is Schur convex, so $x \preceq y$ implies that $f(x) \le f(y)$. Consequently, inequality~\eqref{EqnMaj} implies that
\begin{equation*}
\det(\sigma_x^2 \tilde{U}^\top  D \tilde{U} + \Lambda^{-2}) \le \prod_{i=1}^k (\sigma_x^2 d_i + \ell_i).
\end{equation*}
Further note that equality is clearly achieved when $\tilde{U}$ is a matrix with columns equal to the canonical basis vectors of $\real^p$, which simply picks out the $k$ maximal diagonal entries of $D$ and sorts them according to the diagonal entries of $\Lambda^{-2}$. Finally, we argue that this matrix $\tilde{U}$ corresponds to a choice of the orthogonal matrix $U$. Indeed, define $\hat{W} \in \real^{p \times p}$ to be an orthogonal matrix obtained by taking the first $k$ columns equal to the columns of $W$, and appending with $p-k$ column vectors to obtain an orthonormal basis of $\real^p$. Similarly, define $\hat{U}$ to denote an orthogonal matrix with first $k$ columns equal to the columns of $\tilde{U}$, and the remaining columns chosen to form an orthonormal basis of $\real^p$. If we define $U = \hat{W} \hat{U}^\top $, we have
\begin{equation*}
U^\top  W = \hat{U} \hat{W}^\top  W = \hat{U} \begin{bmatrix} I_k \\ 0_{(p-k) \times k} \end{bmatrix} = \tilde{U},
\end{equation*}
as wanted.

Returning to the objective~\eqref{EqnObjSep} and substituting in the optimal choice of $U$, we obtain the following optimization problem in terms of $D$:
\begin{equation*}
\min_{0 \preceq D \preceq I} \left\{\frac{1}{2} \log \prod_{i=1}^k \left(\frac{\sigma_x^2 d_i}{\ell_i} + 1\right) - \frac{\gamma}{2} \log \prod_{i=1}^p d_i + \frac{\beta}{\sigma_x^2} \sum_{i=1}^p \frac{1}{d_i} \right\}.
\end{equation*}
Clearly, the constraint is equivalent to $0 \le d_i \le 1$ for all $i$. It is easy to see that the optimum is achieved when $d_i = 1$ for all $i > k$. Furthermore, for $i \le k$, the optimal choice of $d_i$ corresponds to minimizing the univariate function
\begin{equation*}
f_i(d) = \frac{1}{2} \log\left(\frac{\sigma_x^2 d}{\ell_i} + 1\right) - \frac{\gamma}{2} \log d + \frac{\beta}{\sigma_x^2 d}
\end{equation*}
over the interval $d \in [0,1]$.

Finally, from the expression~\eqref{EqnT}, we see that the optimal $T$ is then given by
\begin{align*}
T & = \left((\sigma_x^2 I - \Sigma^*) - \frac{1}{\sigma_x^2} (\sigma_x^2 I - \Sigma^*)^2\right)^{-1/2} \frac{\sigma_x^2 I - \Sigma^*}{\sigma_x^2} X + \epsilon \\
& = \left(\sigma_x^2 U(I-D) U^\top  - \sigma_x^2 U(I-D)^2U^\top \right)^{-1/2} U(I-D)U^\top  X + \epsilon \\
& = \frac{1}{\sigma_x} \left(U(D-D^2)U^\top \right)^{-1/2} U(I-D)U^\top  X + \epsilon \\
& = \frac{1}{\sigma_x} U(D-D^2)^{-1/2}U^\top  \cdot U(I-D)U^\top  X + \epsilon \\
& = \frac{1}{\sigma_x} U (D^{-1} - I)^{1/2} U^\top  X + \epsilon.
\end{align*}
We may apply the rotation $U^\top $ to obtain a slightly more attractive form:
\begin{equation*}
T = \frac{1}{\sigma_x} (D^{-1} - I)^{1/2} U^\top  X + \epsilon.
\end{equation*}

In fact, we can simplify this expression even further, since the latter $p-k$ coordinates of $T$ are irrelevant. Indeed, since $d_i = 1$ for $i > k$, we see that the last $p-k$ diagonal entries of $(D^{-1} - I)^{1/2}$ are all 0. Furthermore, if we examine the factor $U^\top  = \hat{U} \hat{W}^\top $, we see that this matrix has first $k$ rows corresponding to a permuted version of the $k$ rows of $W^\top $, and the remaining rows identical to the rows of $\hat{W}^\top $. Multiplication by $(D^{-1} - I)^{1/2}$ results in zeroing out the last $p-k$ rows. Thus, by Lemma~\ref{lemma: invariance}(4), we conclude that we can also express the optimal features in terms of the truncated matrices $D = \diag(d_1, \dots, d_k)$ and $U = W \hat{U}^\top $, as stated in the theorem.

\subsection{Proof of Theorem~\ref{ThmGamZero}}
\label{AppThmGamZero}

If $(T,X,Y)$ are jointly Gaussian with $T = AX + \epsilon$, the problem~\eqref{EqnMutualFish} is equivalent to optimizing
\begin{equation*}
\min_A \left\{\frac{1}{2} \log \det\left(\Sigma_y - \Sigma_{yx} A^\top (A \Sigma_x A^\top  + I)^{-1} A \Sigma_{xy}\right) + \beta \tr(A^\top A)\right\},
\end{equation*}
where we have used Lemma~\ref{LemFeatGauss} to simplify the Fisher information term.
We now make the substitution $B = A\Sigma_x^{1/2}$, so we may rewrite the optimization problem as
\begin{equation*}
\min_B \left\{\frac{1}{2} \log\det\left(\Sigma_y - \Sigma_{yx} \Sigma_x^{-1/2} B^\top  (BB^\top  + I)^{-1} B \Sigma_x^{-1/2} \Sigma_{xy}\right) + \tr\left(\Sigma_x^{-1/2} B^\top B \Sigma_x^{-1/2}\right)\right\}.
\end{equation*}
Furthermore, using the fact that $B^\top (BB^\top  + I)^{-1} B = I - (B^\top B + I)^{-1}$, we may transform the objective into
\begin{equation*}
\min_B \left\{\frac{1}{2} \log \det \left((\Sigma_y - \Sigma_{yx} \Sigma_x^{-1} \Sigma_{xy}) + \Sigma_{yx} \Sigma_x^{-1/2} (B^\top B + I)^{-1} \Sigma_x^{-1/2} \Sigma_{xy}\right) + \beta \tr\left(\Sigma_x^{-1} B^\top B\right)\right\}.
\end{equation*}
Factoring out $(\Sigma_y - \Sigma_{yx} \Sigma_x^{-1} \Sigma_{xy})$ from the first term and then ignoring it (since the term does not depend on $B$), we then obtain
\begin{equation*}
\min_B \left\{\frac{1}{2} \log \det(I + C^\top (B^\top B + I)^{-1} C) + \beta \tr\left(\Sigma_x^{-1} B^\top B\right)\right\},
\end{equation*}
where $C := \Sigma_x^{-1/2} \Sigma_{xy} (\Sigma_y - \Sigma_{yx} \Sigma_x^{-1} \Sigma_{xy})^{-1/2}$.

Let $H = B^\top B + I$. Then the optimization problem further simplifies to
\begin{equation}
\label{EqnHObj}
\min_H \left\{\frac{1}{2} \log \det(I + C^\top  H^{-1} C) + \beta \tr(\Sigma_x^{-1}H)\right\},
\end{equation}
where the optimization is over a restricted set of psd matrices $H$ which may be represented as $B^\top B + I$.

The following lemma shows that the objective~\eqref{EqnHObj} is convex in $H$:

\begin{lemma*}
\label{LemConvex}
The function $\log \det (I+ C^\top  H^{-1} C)$ is convex in $H$.
\end{lemma*}

\begin{proof}
By Sylvester's determinant identity, we have $\det (I_m+A_{m\times n}B_{n\times m}) = \det (I_n +B_{n \times m}A_{m \times n})$. Thus, we have
\begin{align*}
\log \det (I+ C^\top  H^{-1} C) = \log \det (I+ CC^\top  H^{-1} ).
\end{align*}
Using the fact that $\log \det (I + KH^{-1})$ is convex in $H$ for $K \succeq 0$ \cite{DigCov01, Kim15}, we conclude the proof.
\end{proof}

We now take a gradient with respect to $H$ of the objective~\eqref{EqnHObj}. We have the following lemma:

\begin{lemma*}
\label{LemGradient}
The gradient of $\log \det(I + C^\top  H^{-1} C)$ in $H$ is $-H^{-1} C(I + C^\top  H^{-1} C)^{-1} C^\top  H^{-1}$.
\end{lemma*}

\begin{proof}
Consider an $\epsilon$-perturbation of $H$ in a direction $V$ given by $H+\epsilon V$. We shall expand out the function $\log \det(I + C^\top  (H+\epsilon V)^{-1} C)$ and identify the coefficient of $\epsilon$. Note that 
\begin{align*}
(H + \epsilon V)^{-1} &= (H(I + \epsilon H^{-1}V))^{-1}\\
&= (I + \epsilon H^{-1}V)^{-1} H^{-1}\\
&= (I - \epsilon H^{-1}V) H^{-1} + o(\epsilon)\\
&= H^{-1} - \epsilon H^{-1} V H^{-1} + o(\epsilon).
\end{align*}
Hence, we have
\begin{align*}
\log \det(I + C^\top  (H+\epsilon V)^{-1} C) &= \log \det(I + C^\top  (H^{-1} - \epsilon H^{-1} V H^{-1} + o(\epsilon)) C)\\
&= \log \det(I + C^\top  H^{-1} C - \epsilon C^\top  H^{-1} V H^{-1} C + o(\epsilon))\\
&\stackrel{(a)}= -\epsilon \tr((I + C^\top  H^{-1} C)^{-1} C^\top H^{-1} V H^{-1} C) + o(\epsilon)\\
&= -\epsilon \tr(H^{-1}C(I + C^\top  H^{-1} C)^{-1} C^\top  H^{-1}V) + o(\epsilon).
\end{align*}
Here, the equality in $(a)$ follows from the fact $\nabla_X \log \det X = X^{-1}$. Thus, the gradient must be $-H^{-1}C(I + C^\top  H^{-1} C)^{-1} C^\top  H^{-1}$.
\end{proof}

By Lemma~\ref{LemGradient}, we have
\begin{equation*}
\nabla_H \log \det(I + C^\top  H^{-1} C) = -H^{-1} C(I + C^\top  H^{-1} C)^{-1} C^\top  H^{-1}.
\end{equation*}
Since $\nabla_H \tr(\Sigma_x^{-1} H) = \Sigma_x^{-1}$, this means the optimum occurs when
\begin{equation*}
H^{-1} C(I + C^\top  H^{-1} C)^{-1} C^\top  H^{-1} = \beta \Sigma_x^{-1}.
\end{equation*}
If we left-multiply by $C^\top $ and right-multiply by $C$, we obtain
\begin{equation*}
C^\top  H^{-1} C (I + C^\top  H^{-1} C)^{-1} C^\top  H^{-1} C = \beta C^\top  \Sigma_x^{-1} C.
\end{equation*}

We now derive the following useful lemma:

\begin{lemma*}
\label{LemCinvert}
Suppose $M$ is any positive definite matrix. Then $C^\top  M C$ is also invertible.
\end{lemma*}

\begin{proof}
Recall that $\rank(C^\top  M C) = \rank(M^{1/2} C) = \rank(C)$, since $M$ is invertible (cf.\ Chapter 0 of Horn and Johnson~\cite{HorJoh13}). Since $C$ has full column rank by assumption, we conclude that $C^\top  M C$ is full-rank, hence invertible.
\end{proof}

In particular, Lemma~\ref{LemCinvert} implies that both $C^\top  H^{-1} C$ and $C^\top  \Sigma_x^{-1}C$ are invertible. Hence, we can write
\begin{equation}
\label{EqnTomato}
\frac{1}{\beta} (C^\top  \Sigma_x^{-1} C)^{-1} = (C^\top  H^{-1} C)^{-1} (I + C^\top  H^{-1} C) (C^\top  H^{-1} C)^{-1} = (C^\top  H^{-1} C)^{-2} + (C^\top  H^{-1} C)^{-1}.
\end{equation}
Now let $(C^\top  \Sigma_x^{-1} C)^{-1} = UDU^\top $ be the SVD. We know that $U$ is a unitary matrix and $D$ is positive definite. Left-multiplying equation~\eqref{EqnTomato} by $U^\top $ and right-multiplying by $U$ then gives
\begin{equation}
\label{EqnElmo}
\frac{D}{\beta} = U^\top (C^\top  H^{-1} C)^{-2} U + U^\top  (C^\top  H^{-1} C)^{-1} U = (U^\top  C^\top  H^{-1} C U)^{-2} + (U^\top  C^\top  H^{-1} C U)^{-1}.
\end{equation}
Let $\tilde{D}$ be the diagonal matrix which solves $\tilde{D}^{-2} + \tilde{D}^{-1} = \frac{D}{\beta}$. In other words, if $\{\tilde{d}_i\}$ and $\{d_i\}$ are the diagonal entries of $\tilde{D}$ and $D$, respectively, we have
\begin{equation*}
\frac{1}{\tilde{d}_i^2} + \frac{1}{\tilde{d}_i} = \frac{d_i}{\beta},
\end{equation*}
implying that
\begin{equation*}
1 + \tilde{d}_i = \frac{d_i}{\beta} \tilde{d_i}^2,
\end{equation*}
so
\begin{equation*}
\tilde{d}_i = \frac{1 + \sqrt{1 + 4d_i/\beta}}{2d_i/\beta}.
\end{equation*}

It remains to find a proper assignment of $H$, such that $U^\top C^\top H^{-1}CU = \tilde{D}$. Consider the SVD $C = W \Lambda V^\top $. We know that $\Lambda$ is invertible and $V$ is unitary, and that $W^\top W = I_k$. Define $\tilde{W}$ to be the matrix with first columns equal to the columns of $W$, and the remaining columns chosen arbitrarily to form a unitary matrix. Then define $H = \tilde{W} \begin{bmatrix} J & 0 \\ 0 & I \end{bmatrix} \tilde{W}^\top $, where
\begin{equation*}
J := \Lambda V^\top  U \tilde{D}^{-1} U^\top  V \Lambda
\end{equation*}
is full-rank, since it is a product of invertible matrices. We can verify that
\begin{equation*}
H^{-1} = \tilde{W} \begin{bmatrix} \Lambda^{-1} V^\top  U \tilde{D} V \Lambda^{-1} & 0 \\ 0 & I \end{bmatrix} \tilde{W}^\top ,
\end{equation*}
and
\begin{equation*}
U^\top  C^\top  H^{-1} C U = U^\top  V \Lambda W^\top  \tilde{W} \begin{bmatrix} \Lambda^{-1} V^\top  U \tilde{D} V \Lambda^{-1} & 0 \\ 0 & I \end{bmatrix} \tilde{W}^\top  W \Lambda V^\top  U = \tilde{D}.
\end{equation*}
Thus, equation~\eqref{EqnElmo} is satisfied.

It remains to choose $B$ (and $A = B \Sigma_x^{-1/2}$) appropriately. Note that
\begin{equation*}
\tilde{W}^\top  B^\top B \tilde{W} = \tilde{W}^\top  H \tilde{W} - I = \begin{bmatrix} J-I & 0 \\ 0 & 0 \end{bmatrix}.
\end{equation*}
If we write the SVD $J-I = S \Gamma S^\top $, assuming $J - I \succeq 0$, and let $\tilde{S} \in \real^{p \times k}$ be the matrix formed by taking $S \in \real^{k \times k}$ and zero-filling the remaining $p-k$ rows, we can take $B = \Gamma^{1/2} \tilde{S}^\top  \tilde{W}^\top  = \Gamma^{1/2} S^\top  W^\top $. Then
\begin{equation*}
\tilde{W}^\top  B^\top  B \tilde{W} = \tilde{S} \Gamma \tilde{S^\top } = \begin{bmatrix} J-I & 0 \\ 0 & 0 \end{bmatrix},
\end{equation*}
which is what we want.

Lastly, note that $J - I \succeq 0$ holds when $\beta$ is sufficiently small (as a function of $(\Sigma_x, \Sigma_y, \Sigma_{xy})$). Indeed, we have
\begin{equation*}
\frac{1}{\tilde{d}_i} = \frac{2d_i/\beta}{1+\sqrt{1 + 4d_i/\beta}} \ge \frac{2d_i/\beta}{3 \sqrt{d_i/\beta}} = \frac{2}{3} \sqrt{\frac{d_i}{\beta}}
\end{equation*}
if $\beta < \frac{d_i}{2}$, so $\tilde{D}^{-1} \succeq \frac{2}{3} \sqrt{\frac{d_{\min}}{\beta}} I$, where $d_{\min} := \min_i d_i$, implying that
\begin{equation*}
J \succeq \Lambda V^\top  U \cdot \frac{2}{3} \sqrt{\frac{d_{\min}}{\beta}} I \cdot U^\top  V \Lambda = \frac{2}{3} \sqrt{\frac{d_{\min}}{\beta}} \Lambda^2.
\end{equation*}
Hence,
\begin{equation*}
\lambda_{\min}(J) \ge \frac{2}{3} \sqrt{\frac{d_{\min}}{\beta}} \lambda_{\min}^2(\Lambda),
\end{equation*}
and we can guarantee that the final expression is at least 1 by making $\beta$ sufficiently small.

\subsection{Proof of Theorem~\ref{ThmMMSEId}}
\label{AppThmMMSEId}

Suppose $T = AX + \epsilon$ with $\epsilon \sim N(0, I)$ and $A \in \real^{k \times p}$, and denote the subblocks of the covariance matrix of $(X,Y)$ by $\Sigma_x$, $\Sigma_y$, and $\Sigma_{xy}$.

Using standard formulas for jointly Gaussian variables, we have
\begin{align*}
\Cov(Y|T) & = \Sigma_y - \Sigma_{yt} \Sigma_{t}^{-1} \Sigma_{ty} = \Sigma_y - \Sigma_{yx} (A\Sigma_x A^\top  + I)^{-1} \Sigma_{xy}.
\end{align*}
Furthermore, we have $\mmse(Y|T) = \tr(\Cov(Y|T))$. In addition, Lemma~\ref{LemFeatGauss} implies that $\Phi(X|T) = \tr(A^\top A)$.
Hence, we can simplify the objective to
\begin{equation*}
\min_A \left\{-\tr\left(\Sigma_{yx} A^\top  (A \Sigma_x A^\top  + I)^{-1} A \Sigma_{xy}\right) + \beta \tr(A^\top A)\right\}.
\end{equation*}

Now let $B = A\Sigma_x^{1/2}$. The objective then becomes
\begin{equation}
\label{EqnBObj}
\min_B \left\{-\tr\left(\Sigma_{yx} \Sigma_x^{-1/2} B^\top  (BB^\top  + I)^{-1} B \Sigma_x^{-1/2} \Sigma_{xy}\right) + \beta \tr\left(\Sigma_x^{-1/2} B^\top B \Sigma_x^{-1/2}\right)\right\}.
\end{equation}
Also note that by the Woodbury matrix formula, we can write
\begin{equation*}
(B^\top B + I)^{-1} = I - B^\top (I + BB^\top )^{-1} B,
\end{equation*}
so that the objective~\eqref{EqnBObj} is equal to
\begin{equation*}
-\tr\left(\Sigma_{yx} \Sigma_x^{-1/2}\left(I - (B^\top B + I)^{-1}\right) \Sigma_x^{-1/2} \Sigma_{xy}\right) + \beta\tr\left(\Sigma_x^{-1/2} B^\top B \Sigma_x^{-1/2}\right).
\end{equation*}
In particular,
the optimization problem~\eqref{EqnBObj} may be viewed as an optimization problem over $B^\top B$.
We write $B^\top B = U DU^\top $, where $D \in \real^{p \times p}$ is a diagonal psd matrix (possibly noninvertible) and $U \in \real^{p \times p}$ satisfies $U^\top U = I = UU^\top $. We may then equivalently optimize
\begin{align*}
& \min_B \left\{\tr\left(\Sigma_{yx} \Sigma_x^{-1/2}(B^\top B + I)^{-1} \Sigma_x^{-1/2} \Sigma_{xy}\right) + \beta \tr\left(\Sigma_x^{-1/2} B^\top B \Sigma_x^{-1/2}\right)\right\} \\
& = \min_{D,U} \left\{-\tr\left(\Sigma_{yx} \Sigma_x^{-1/2} U(I + D)^{-1}U^\top  \Sigma_x^{-1/2} \Sigma_{xy}\right) + \beta \tr\left(\Sigma_x^{-1/2} UDU^\top  \Sigma_x^{-1/2}\right) \right\}.
\end{align*}
A final transformation, using the cyclic property of trace, then gives
\begin{equation}
\label{EqnGenObj}
\min_{D, U} \left\{\tr\left((D+I)^{-1} \cdot U^\top  \Sigma_x^{-1/2} \Sigma_{xy} \Sigma_{yx} \Sigma_x^{-1/2} U\right) + \tr\left(D \cdot \beta U^\top  \Sigma_x^{-1} U\right)\right\}.
\end{equation}

Plugging in $\Sigma_x = \sigma_x^2I$, we need to solve
\begin{equation}
\label{EqnCovId}
\min_{D,U} \left\{\tr\left((D+I)^{-1} \cdot U^\top  \Sigma_{xy} \Sigma_{yx} U\right) + \tr(\beta D)\right\}.
\end{equation}
We claim that the optimum $U$ for a fixed $D$ does not depend on the value of $D$. Indeed, suppose WLOG that $D = \diag(d_1, \dots, d_p)$, where $d_1 \le \cdots \le d_p$. Applying the lower bound of Lemma~\ref{LemTrace} with $A = (D+I)^{-1}$ and $B = U^\top  \Sigma_{xy} \Sigma_{yx} U$, we see that
\begin{equation}
\label{EqnLower}
\tr\left((D+I)^{-1} \cdot U^\top  \Sigma_{xy} \Sigma_{yx} U\right) \ge \sum_{i=1}^p \frac{1}{d_i + 1} \cdot \lambda_i,
\end{equation}
where $\lambda_1 \le \cdots \le \lambda_p$ are the eigenvalues of $U^\top  \Sigma_{xy} \Sigma_{yx} U$. However, the eigenvalues $\{\lambda_i\}$ are also the eigenvalues of $\Sigma_{xy} \Sigma_{yx}$, which do not depend on $U$. Furthermore, the lower bound~\eqref{EqnLower} is achieved when the columns of $U$ are the eigenvectors of $\Sigma_{xy} \Sigma_{yx}$, in increasing order from left to right according to the values of the $\lambda_i$'s.

We now turn to optimizing with respect to $D$.
Returning to the objective~\eqref{EqnCovId}, we need to minimize
\begin{equation*}
\min_{d_1, \dots, d_p} \sum_{i=1}^p \left(\frac{1}{d_i + 1} \cdot \lambda_i + \beta d_i\right).
\end{equation*}
Note that we may optimize each term separately; setting $f(d) = \frac{\lambda}{d+1} + \beta d$, we see that $f'(d) = \frac{-\lambda}{(d+1)^2} + \beta$, so taking $f'(d) = 0$ gives $d = \sqrt{\frac{\lambda}{\beta}} - 1$. Thus, we conclude that the optimum is achieved when
\begin{equation}
\label{EqnDvals}
d_i =
\begin{cases}
\sqrt{\frac{\lambda_i}{\beta}} - 1 & \text{ if } \lambda_i \ge \beta, \\
0 & \text{otherwise.}
\end{cases}
\end{equation}

We now need to transform this joint solution $(U,D)$ into a choice of the matrices $B$ and $A$. If we construct $B$ to be $D^{1/2}U^\top $, where $D$ is the $p \times p$ matrix with values given by equation~\eqref{EqnDvals}, then we will have $B^\top B = UDU^\top $. We may then take
$A = B \Sigma_x^{-1/2} = \frac{1}{\sigma_x}D^{1/2} U^\top $. Lastly, note that by Lemma~\ref{lemma: invariance}(4), we may safely ignore the entries of $D$ and $U$ corresponding to the zero eigenvalues, leading to the statement of the theorem.

\section{Properties of Fisher information}
\label{AppFisher}

For the results in this appendix, we will assume all random variables have densities that are twice differentiable and have tails that decay sufficiently quickly. For instance, this happens when the random variables under consideration are preprocessed by convolving with a small amount of Gaussian noise. The precise technical conditions, which may be found in Lehmann and Casela \cite[Lemma 5.3, pg. 116]{Leh06}, essentially allow for interchanging of the integral and differential operators.

We begin with a few definitions:

\begin{definition}
The \emph{Fisher information} of $X$ is
$$J(X) = J(p_X) = \int_{\real^d} p_X(x) \|_2\nabla \log p_X(x)\|^2 \, dx = \E\left[\|\nabla \log p_X(X)\|_2^2\right].$$
\end{definition}

\begin{definition}
The \emph{Fisher information matrix} of $X$ is 
\begin{align*}
\J(X) = \J(p_X) &\,=\, \int_{\real^d} p_X(x) (\nabla \log p_X(x))(\nabla \log p_X(x))^\top dx
\,=\, -\int_{\real^d} p_X(x) \nabla^2 \log p_X(x) \, dx,
\end{align*}
so that $J(X) = \tr(\J(X))$.
\end{definition}

\begin{definition}
The \emph{mutual Fisher information} is the mutual version of the Fisher information:
$$J(X;Y) = J(Y\,|\,X) - J(Y),$$
where $J(Y|X) = \int J(Y | X = x) p_X(x) dx$.
\end{definition}

\begin{definition}
The \emph{mutual Fisher information matrix} is the mutual version of the Fisher information matrix:
$$\J(X;Y) = \J(Y\,|\,X) - \J(Y),$$
so that $J(X;Y) = \tr(\J(X;Y))$.
\end{definition}

\begin{definition}
The \emph{statistical Fisher information} of $X$ with respect to $Y$ is
$$\Phi(X\,|\,Y) = \int_{\real^n} p_Y(y) \int_{\real^n} p_{X|Y}(x\,|\,y) \|\nabla_y \log p_{X|Y}(x\,|\,y)\|_2^2 dx \, dy.$$
\end{definition}

\begin{definition}
The \emph{statistical Fisher information matrix} of $X$ with respect to $Y$ is
\begin{align*}
\mathbf{\Phi}(X\,|\,Y) &= \int_{\real^n} p_Y(y) \int_{\real^n} p_{X|Y}(x\,|\,y) (\nabla_y \log p_{X|Y}(x\,|\,y))(\nabla_y \log p_{X|Y}(x\,|\,y))^\top dx \, dy \\
&= -\int_{\real^n} p_Y(y) \int_{\real^n} p_{X|Y}(x\,|\,y) \nabla^2_y \log p_{X|Y}(x\,|\,y) dx \, dy,
\end{align*}
so that $\Phi(X\,|\,Y) = \tr(\mathbf{\Phi}(X\,|\,Y))$.
\end{definition}

\begin{lemma*}
\label{Lem:JPhi}
Under mild regularity conditions, the random variables $(X, Y)$ satisfy the identities
\begin{align*}
J(X;Y) &= \Phi(X\,|\,Y), \\
\J(X;Y) &= \mathbf{\Phi}(X\,|\,Y).
\end{align*}
\end{lemma*}

\begin{proof}
From the factorization
$$p_X(x) p_{Y|X}(y\,|\,x) = p_{XY}(x, y) = p_Y(y) p_{X|Y}(x\,|\,y),$$
we have
\begin{align*}
-\nabla^2_y \log p_{Y|X}(y\,|\,x) = -\nabla^2_y \log p_Y(y) - \nabla^2_y \log p_{X|Y}(x\,|\,y),
\end{align*}
where $\nabla^2_y$ denotes the Hessian matrix of second derivatives with respect to $y$.
We integrate both sides with respect to $p_{XY}(x, y)$.
On the left-hand side, we obtain the expected Fisher information matrix $\J(Y\,|\,X)$.
The first term on the right-hand side gives $\J(Y)$, while the second term gives $\mathbf{\Phi}(X\,|\,Y)$.
Therefore,
$$\J(Y\,|\,X) = \J(Y) + \mathbf{\Phi}(X\,|\,Y),$$
so that
$$\J(X;Y) = \J(Y\,|\,X) - \J(Y) = \mathbf{\Phi}(X\,|\,Y).$$
Taking the trace then gives
\begin{equation*}
J(X;Y) = \tr(\J(X;Y)) = \tr(\mathbf{\Phi}(X\,|\,Y)) = \Phi(X;Y).
\end{equation*}
\end{proof}

\begin{lemma*}\label{Lem:JConvex}
The Fisher information function $J$ is convex; i.e., for any distributions $p$ and $q$ on $\real^n$ and any $\lambda \in (0,1)$, we have
\begin{align*}
J(\lambda p + (1-\lambda)q) \leq \lambda J(p) + (1-\lambda) J(q).
\end{align*}
Furthermore, equality holds in the above inequality if and only if $p$ and $q$ are identical. 
\end{lemma*}

\begin{proof}
Our proof is adapted from Bobkov et al.~\cite{BobEtal14}. Consider the function from $\real^{n} \times \real_+ \to \real$ given by $g(u, v) := g(u_1, u_2, \dots, u_n, v) = \sum_{i=1}^n \frac{u_i^2}{v}$. It is easy to check that the Hessian of this function is
$$\frac{2}{v} \cdot
\begin{pmatrix}
I &a\\
a^\top  &b
\end{pmatrix}
,$$
where $b = \sum_{i=1}^n \frac{u_i^2}{v^2}$ and $a = \left(\frac{-u_1}{v}, \dots, \frac{-u_n}{v}\right)^\top $. Using Schur complements, we see that the Hessian is positive semi-definite for all points in $\real^{n} \times \real_+$. Furthermore, the (unique) direction of zero curvature at $(u,v)$ is seen to be $(u,v)$. Using the convexity of $g$ and considering the points $(\nabla p, p)$ and $(\nabla q, q)$, we see that 
\begin{align*}
\frac{\| \lambda \nabla p(x) + (1-\lambda) \nabla q(x) \|_2^2}{\lambda \nabla p(x) + (1-\lambda) \nabla q(x)} \leq \lambda \frac{\| \nabla p(x) \|_2^2}{p(x)} + (1-\lambda) \frac{\| \nabla q(x) \|_2^2}{q(x)}.
\end{align*}
Integrating the above inequality over all $x$, we conclude that
\begin{align*}
J(\lambda p + (1-\lambda)q) \leq \lambda J(p) + (1-\lambda) J(q).
\end{align*}
Equality holds if and only if for every $x \in \real^n$, we have $(\nabla p, p) \propto (\nabla q, q)$, which is only possible if $p$ and $q$ are identical.
\end{proof}

\begin{corollary*}
We have the inequality $J(X;Y) \geq 0$. Furthermore, $J(X;Y) =0$ if and only if $X \ind Y$.
\end{corollary*}
\begin{proof}
By convexity of $J$, we have $J(Y|X) \geq J(Y)$, or equivalently, $J(X; Y) \geq 0$.
\end{proof}

\begin{corollary*}[Data processing inequality]
If $\Theta \to X \to Y$ is a Markov chain, then $J(\Theta | X) \geq J(\Theta | Y)$.
\end{corollary*}
\begin{proof}
By the above corollary, we have $J(\Theta | X,Y) \geq J(\Theta | Y)$. Using the Markov chain $\Theta \to X \to Y$ gives $J(\Theta | X, Y) = J(\Theta | X)$, completing the proof. 
\end{proof}

\begin{remark*}
Convexity of $J(\cdot)$ is essentially identical to the data-processing properties of Fisher information. It is worth remarking that Zamir \cite{Zam98} established a chain rule for the statistical Fisher information, as follows:
\begin{align*}
\mathbf \Phi(X, Y | \Theta) = \mathbf \Phi(X | \Theta) + \mathbf \Phi(Y|\Theta \| X),
\end{align*}
where $\mathbf \Phi(Y|\Theta \| X)$ is interpreted as the average $\mathbf \Phi(p_{Y|X=x} | \Theta)$ over the marginal of $X$. This also implies that if $\Theta \to X \to Y$, then $\mathbf \Phi(X, Y | \Theta) = \mathbf \Phi(X | \Theta) = \mathbf \Phi(Y | \Theta) + \mathbf \Phi(X | \Theta \| Y)$, so that $\mathbf \Phi(X | \Theta) \geq \mathbf \Phi(Y | \Theta)$. Stated differently, $J(\Theta | X) \geq J(\Theta | Y)$.
\end{remark*}

\begin{lemma*}
Let $X$ and $Y$ be random variables taking values on $\real^m$ and $\real^n$, respectively. Then $J(X,Y) = J(X|Y) + J(Y|X)$.
\end{lemma*}

\begin{proof}
The Fisher information matrix $\tilde J(X,Y)$ is given by
\begin{align*}
\J(X, Y) &\,=\, -\int_{\real^{m+n}} p_{XY}(x, y) \nabla^2 \log p_{XY}(x, y) \, dxdy.
\end{align*}
Note that the trace of $\J(X,Y)$ is the same as the sum of the traces of
\begin{equation*}
\J(X,Y)_{1:m, 1:m} \qquad \text{and} \qquad \J(X,Y)_{m+1:m+n, m+1:m+n}.
\end{equation*}
To evaluate the first quantity, we factorize $p_{XY}(x, y) = p_Y(y)p_{X|Y}(x|y)$ and notice that the Hessian involves derivatives only in $x$. Thus, $\nabla_x^2 \log p_{XY}(x, y) = \nabla_x^2 \log p_{X|Y}(x|y)$. It is now easy to check that
\begin{equation*}
\tr(\J(X,Y)_{1:m, 1:m}) = J(X|Y).
\end{equation*}
A similar argument shows that
\begin{equation*}
\J(X,Y)_{m+1:m+n, m+1:m+n} = J(Y|X).
\end{equation*}
\end{proof}

\section{Linear algebraic lemmas}
\label{SecAux}

Finally, we collect several linear algebraic lemmas that are used in our derivations. Throughout this section, let $M_n$ denote the space of $n \times n$ matrices with real-valued entries.

\begin{lemma*} [Fan inequality, Theorem 4.3.47 in Horn \& Johnson~\cite{HorJoh13}]
\label{LemFan}
Let $A, B \in M_n$ be Hermitian. Let $\lambda(A), \lambda(B)$, and $\lambda(A+B)$ denote the real $n$-vectors of eigenvalues of $A, B$, and $A+B$, respectively. Then $\lambda(A)^{\downarrow} + \lambda(B)^{\downarrow}$ majorizes $\lambda(A+B)$.
\end{lemma*}

\begin{lemma*} [Theorem 4.3.53 in Horn \& Johnson~\cite{HorJoh13}]
\label{LemTrace}
Let $A, B \in M_n$ be Hermitian and have respective vectors of eigenvalues $\lambda(A) = \{\lambda_i(A)\}_{i=1}^n$ and $\lambda(B) = \{\lambda_i(B)\}_{i=1}^n$. Then
\begin{equation*}
\sum_{i=1}^n \lambda_i(A)^{\downarrow} \lambda_i(B)^{\uparrow} \le \tr(AB) \le \sum_{i=1}^n \lambda_i(A)^{\downarrow} \lambda_i(B)^{\downarrow}.
\end{equation*}
\end{lemma*}

\begin{lemma*} [Corollary 4.3.39 in Horn \& Johnson~\cite{HorJoh13}]
\label{LemMaxEigs}
Let $A \in M_n$ be Hermitian with eigenvalues $\lambda_1(A) \ge \lambda_2(A) \ge \cdots \ge \lambda_n(A)$  and suppose $1 \le m \le n$. Then
\begin{equation*}
\lambda_1(A) + \cdots + \lambda_m(A) = \max_{V \in M_{n,m}: V^\top V = I_m} \tr(V^\top  AV).
\end{equation*}
\end{lemma*}

\end{document}